\documentclass[journal]{IEEEtran}

\makeatletter
\newcommand{\rmnum}[1]{\romannumeral #1}
\newcommand{\Rmnum}[1]{\expandafter\@slowromancap\romannumeral #1@}
\makeatother

\usepackage{multirow}
\usepackage{diagbox}
\usepackage{multirow}
\usepackage{amsmath}
\usepackage{amsfonts}
\usepackage{amssymb}
\usepackage{fancyhdr}
\usepackage{mathtools}
\usepackage{amsthm}
\usepackage{array}
\usepackage{booktabs} 
\usepackage{graphicx}
\usepackage{float}
\usepackage{enumerate}
\usepackage{type1cm}
\usepackage{bbm}
\usepackage{subfigure}
\usepackage{algorithm}
\usepackage{algorithmic}
\usepackage{booktabs}
\usepackage{verbatim}
\usepackage{color}
\usepackage[colorlinks=True,
linkcolor=red,
anchorcolor=green,
citecolor=blue
]{hyperref}

\newfloat{figtab}{htb}{fgtb}
\makeatletter
\newcommand\figcaption{\def\@captype{figure}\caption}
\newcommand\tabcaption{\def\@captype{table}\caption}
\makeatother

\ifCLASSINFOpdf
\else
\fi
\newtheorem{Theorem}{\bf Theorem}[section]
\newtheorem{Definition}{\bf Definition}[section]
\newtheorem{Remark}{\bf Remark}[section]

\newtheorem{Proposition}{\bf Proposition}[section]

\begin{document}
	%
	\title{Multi-Tensor Network Representation for High-Order Tensor Completion}
	%
	%
	%
	
	\author{Chang Nie$^{\href{https://orcid.org/0000-0001-9369-4096}{\includegraphics[scale=0.06]{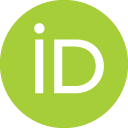}}}$, Huan Wang$^{\href{https://orcid.org/0000-0003-2563-1135}{\includegraphics[scale=0.06]{fig/ORCIDiD_1.png}}}$, and Zhihui Lai$^{\href{https://orcid.org/0000-0002-4388-3080}{\includegraphics[scale=0.06]{fig/ORCIDiD_1.png}}}$
	\thanks{Manuscript received September 22, 2021. This work was supported in part by the Postgraduate Research \& Practice Innovation Program (KYCX21\_0304) of Jiangsu Province, China, and in part by the National Science Fund of China under Grant 61703209. (Corresponding author: Huan Wang.)}
		\thanks{Chang Nie and Huan Wang are with the School of Computer Science and Engineering, Nanjing University of Science and Technology (NJUST), Nanjing 210014, China (e-mail: changnie$@$njust.edu.cn, wanghuanphd$@$njust.edu.cn).}
		\thanks{Zhihui Lai is with the College of Computer Science and Software Engineering, Shenzhen University, Shenzhen 518055, China also with the Laboratory of Intelligent Information Processing, Shenzhen University, Shenzhen 518060, China, and also with the Institute of Textiles and Clothing, Hong Kong Polytechnic University, Hong Kong (e-mail: lai\_zhi\_hui$@$163.com).
		}
	}
	
	%
	%

	\markboth{$ $}
	{Nie \MakeLowercase{\textit{et al.}}: Multi-Tensor Network Representation for High-Order Tensor Completion}
	%


	
	\maketitle
	
	\begin{abstract}
		This work studies the problem of high-dimensional data (referred to as tensors) completion from partially observed samplings. We consider that a tensor is a superposition of multiple low-rank components. In particular, each component can be represented as multilinear connections over several latent factors and naturally mapped to a specific tensor network (TN) topology. In this paper, we propose a fundamental tensor decomposition (TD) framework: Multi-Tensor Network Representation (MTNR), which can be regarded as a linear combination of a range of TD models, e.g., CANDECOMP/PARAFAC (CP) decomposition, Tensor Train (TT), and Tensor Ring (TR). Specifically, MTNR represents a high-order tensor as the addition of multiple TN models, and the topology of each TN is automatically generated instead of manually pre-designed. For the optimization phase, an adaptive topology learning (ATL) algorithm is presented to obtain latent factors of each TN based on a rank incremental strategy and a projection error measurement strategy. In addition, we theoretically establish the fundamental multilinear operations for the tensors with TN representation, and reveal the structural transformation of MTNR to a single TN. Finally, MTNR is applied to a typical task, tensor completion, and two effective algorithms are proposed for the exact recovery of incomplete data based on the Alternating Least Squares (ALS) scheme and Alternating Direction Method of Multiplier (ADMM) framework. Extensive numerical experiments on synthetic data and real-world datasets demonstrate the effectiveness of MTNR compared with the start-of-the-art methods.

	\end{abstract}
	
	%
	
	\IEEEpeerreviewmaketitle
	
	\section{Introduction}
	%
	%
	%
	%
	\IEEEPARstart{W}{ith} the rapid development of sensing and information technology in recent years, multimodal and large-scale data displayed in the form of a multi-dimensional array (referred to as tensors) are ubiquitous in various types, e.g., color images~\cite{ref1_1, ref1}, videos~\cite{ref3}, and seismic data~\cite{zhu2016seismic}, etc. Tensors provide a favorable data structure and widely utilized in data processing and analysis, spanning from the fields of computer vision~\cite{ref4,ref5}, machine learning~\cite{ref6,ref7}, dimension compression~\cite{ref8}, to signal processing~\cite{ref9,ref10}. Unfortunately, missing entry and noise contamination problems are prevalent during data acquisition and transformation, leading to local detail incompleteness and global structure sparseness of the obtained results. That severely limits the practical value and causes the performance degradation of related applications~\cite{ref11}. The estimation of missing elements based on the inherent structure information of the observed entries is termed as tensor completion (TC)~\cite{ref12,ref13,ref14,ref15}.
	
	The theoretical foundation of TC is the low-rank representation framework~\cite{ref11}, which captures the global information by exploiting the low-rank property of incomplete data. Mathematically, the corresponding low-rank TC (LRTC) problem can be derived into the following optimization model as
	\begin{equation}\begin{split}
			\min_{\boldsymbol{\mathcal{X}}}\ \ rank(\boldsymbol{\mathcal{X}})\quad\text{s.t.}\ \  P_\Omega(\boldsymbol{\mathcal{X}})= P_\Omega(\boldsymbol{\mathcal{M}}),
			\label{eq1}
	\end{split}\end{equation}
	where $\mathcal{M}$ is the observed incomplete tensor and $P_\Omega$ is a linear operator that projects the elements in the observing entries set $\Omega$ to itself and others to zeros. 
	
	For the 2D cases, the matrix rank is widely adopted to describe the degree of low-rankness of the data~\cite{ref16}. However, the rank function is nonconvex and hard to optimize. Since the nuclear norm is the tightest convex envelope of the rank of a matrix, which is frequently applied in the low-rank model~\cite{ref12, ref17} and addressed by the singular value threshold (SVT)~\cite{ref45} algorithm. For the matrix $M\in \mathbb{R} ^{n\times n}$ of rank r with $O(nr\log^2 n)$ observed entries satisfying certain incoherence conditions~\cite{ref18}, the missing elements will be recovered exactly with high probability. Nevertheless, the recovered matrix based on nuclear norm minimization strategy is usually suboptimal. Recently, several nonconvex continuous surrogate functions are employed to approximate the rank function, e.g., $L_p$ norm~\cite{ref19}, Capped $L_1$ norm~\cite{ref20}, and have achieved better estimation in the assist of nonconvex optimization techniques~\cite{bolte2014proximal, lu2016nonconvex}.
	
	For the N-D cases, the high-order LRTC problem is challenging due to the flexibility and diversity of tensor algebra operations and inconsistent definition of tensor rank~\cite{ref1_1}. In the multilinear products algebraic framework~\cite{ref29}, multilinear tensor rank~\cite{ref1, ref5} is defined as a vector consisting of the ranks of all mode unfolding matrices. In this context, the Sum of Nuclear Norm (SNN)~\cite{ref5, ref22} is extensively utilized as a relaxed convex surrogate of tensor rank. The relaxed convex formulation of (\ref{eq1}) can be written as
	\begin{equation}\begin{split}
			\min_{\boldsymbol{\mathcal{X}}}\ \ \sum_{i=1}^N \alpha_i||\boldsymbol{{X}}_{(i)}||_* \quad \text{s.t.}\ \  P_\Omega(\boldsymbol{\mathcal{X}})= P_\Omega(\boldsymbol{\mathcal{M}}),
			\label{eq2}
	\end{split}\end{equation}
	where $\boldsymbol{X}_{(i)}$ represents the mode-$i$ unfolding matrix of $\boldsymbol{\mathcal{X}}$ and the $||\cdot||_*$ is nuclear norm defined as the sum of singular values, e.g., $||\boldsymbol{A}||_*=\sum_{i}\sigma_i(\boldsymbol{A})$. However, the unfolding operation results in the loss of structural information, and SNN is not the tightest convex relaxation of multilinear tensor rank~\cite{ref23}. In the algebra framework of tensor-product (t-product)~\cite{ref24}, third-order tensors can be regarded as extended linear operators of matrices. Kilmer et al.~\cite{ref25} introduce the definition of tensor tubal rank based on the tensor singular value decomposition(t-SVD)~\cite{ref24}, which accurately estimated the low-rank structure of 3th-order tensors with a complete theoretical guarantee. Tensor nuclear norm (TNN)~\cite{ref2}, a convex surrogate of tensor tubal rank, has a tight recovery bound and is applied for LRTC. Meantime, t-products with invertible transform can be extended to higher dimensions directly~\cite{ref26,ref27} and implemented in other domains~\cite{ref28}. 
	
	The huge advantage of the aforementioned rank-minimization-based TC methods is to estimate the tensor rank implicitly without specifying it in advance. Nevertheless, all existing approaches require a lot of singular value decomposition (SVD)\cite{ek2012singular} calculations in the optimization stage, which is computationally prohibitive and not applicable for large-scale scenariosa~\cite{ref44}. Another promising direction to address (\ref{eq1}) effectively is tensor decomposition (TD)~\cite{ref29,ref30,ref31}, which captures the intrinsic global structure of the data on latent space with multilinear operations over a set of latent factors. The TD-based LRTC model can formulated as
	\begin{equation}\begin{split}
			\min_{[\mathcal{G}]}\ \ ||P_\Omega(\Re([\mathcal{G}])) - P_\Omega(\mathcal{M})||_F^2,
			\label{eq3}
	\end{split}\end{equation}
	here $\Re([\mathcal{G}])$ denotes the recovered tensor generated by a set of latent factors $[\mathcal{G}]$. Two of the most classic and universally applied TD models are CANDECOMP/PARAFAC (CP) decomposition~\cite{ref32} and Tucker decomposition~\cite{ref33}. CP decomposition~\cite{ref32} represents an N-order tensor as a sum of rank-1 tensor factors and defines the CP rank as the smallest number of factors to eliminate structural loss. Tucker decomposition~\cite{ref33} decomposes a tensor as multilinear operations among a core tensor and multiple factor matrix. \textit{Tensor network} (TN)~\cite{ref34, ref35}, one of the powerful tools in physics, is broadly applied to solving large-scale optimization problems and provides graphical computation interpretation. TD models based on TN representation has recently attracted much attention, including tensor train (TT)~\cite{ref36} representation and tensor ring (TR)~\cite{ref37} representation, which can overcome the Curse of Dimensionality~\cite{ref34} with linear storage cost $\mathcal{O}(NIR^2)$. Lately, TN is exploited to explore more complex topology structures~\cite{ref36,ref37,ref47} for TD, e.g., Projected Entangled Pair States (PEPS)~\cite{ref38} and Fully-Connected TN (FCTN)~\cite{ref39}.
	
\begin{figure*}[htbp]
		\centering
		\includegraphics[width=.8\textwidth]{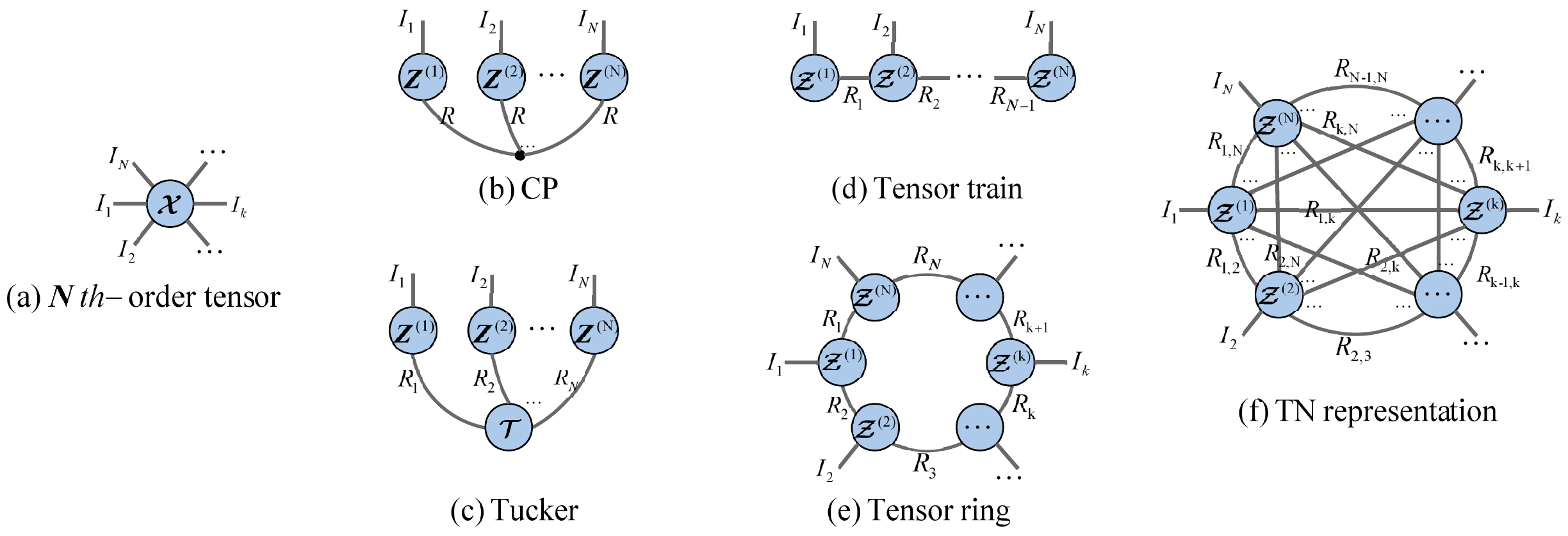} 
		\caption{TN graphical representations. (a) An $N$-order tensor, (b) CP decomposition, where black dot denotes a hyperedge, (c) Tucker decomposition, (d) tensor train (TT) decomposition, (e) tensor ring (TR) decomposition, (f) standard tensor network representation. Note that the topology of the standard tensor network representation can be viewed as a complete graph before removing all the rank-1 edges.}
		\label{img2}
	\end{figure*}

	
	However, most of the available high-dimensional data suffer the difficulties of mode-dimension unbalance~\cite{ref40} (e.g., filters in convolution neural networks), and mode-correlation discrepancy (e.g., spatial modes in color image are strongly intercorrelated but their weakly correlated with channel mode). Such problems lead to the limited representation and flexibility of TD models with fixed topology. Thus, the following two limitations must be considered: (\rmnum{1}) model selection and (\rmnum{2}) rank determination. Notice that the critical difference among various TD models lies in the latent factors connections (also called multilinear operations), as shown in Fig.\ref{img2}. Thus, a current encouraging direction is to explore the data-independent decomposition topology by virtue of TN\cite{ref47, ref49}. In~\cite{ref47}, Hashemizadeh et al. gradually update the TD topology structure based on greedy search and rank incremental strategy. However, the selection of promising edges is tiring since it requires traversing and iterating all possible options. The genetic algorithm is used in~\cite{ref49} by Li and Sun to determine the connection relationship of factors. Since the rank determination for a fixed topology is NP-hard~\cite{ref21}. Cheng et al. ~\cite{ref50} adopt reinforcement learning (RL) to determine the rank of TR representation and applied it to neural network compression tasks. In~\cite{ref48}, the TR rank is gradually increased based on the measured sensitivity of the approximation error of factor. Also, the sparsity-inducing prior is employed by authors in~\cite{ref51, ref52} to determining the optimal rank for TD models. Although many TD-based approaches exhibit favorable characteristics, they are either computationally prohibitive~\cite{ref39, ref47, ref49} or with limited generality~\cite{ref48, ref52, ref60}.
	
	According to the subspace clustering theory~\cite{ref53}, usually the high-dimensional data, e.g., images and videos, which are located close to low-order structures related to several classes to which the data belong, and this property are widely used in unsupervised learning tasks include subspace clustering ~\cite{ref42}, infrared small target detection~\cite{ref41}, latent convex tensor decomposition~\cite{ref43}. In this paper, we present a Multi-Tensor Network Representation (MTNR) framework for high-order tensor decomposition and apply it to the LRTC problem. Our goal is to decompose high-dimensional data into multiple low-rank components, and each component with TN representation can be represented as multilinear connections over several latent factors and naturally mapped to a specific topology. In particular, the structure of MTNR is explored adaptively from the instance data and can selectively establish connections with appropriate intensity for any two factors to capture correlations between data modes. To the best of our knowledge, this work is the first to utilize a combination of multiple TN models for high-order tensor decomposition and completion. In summary, the contribution of this article is threefold.
	
	\begin{figure*}[ht]
		\centering
		\includegraphics[width=.8\textwidth]{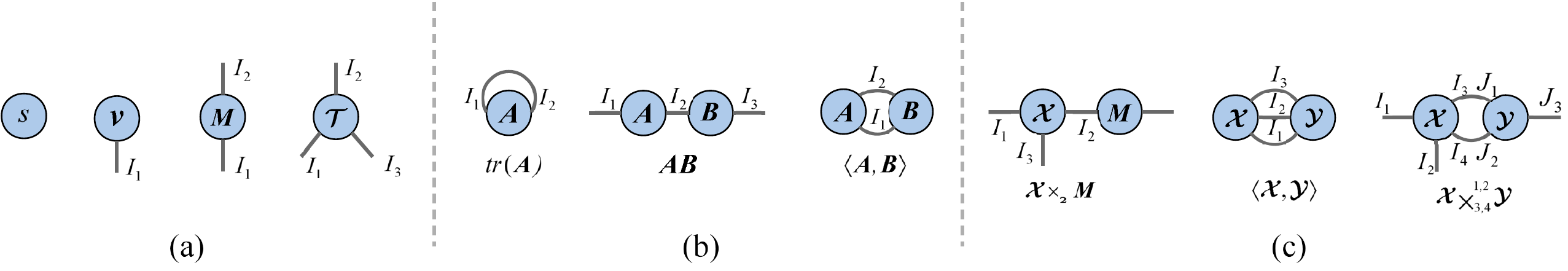}
		\caption{Tensor diagrams. (a) The graphical representation of a scalar $s\in \mathbb{R}$, a vector $\boldsymbol{v}\in \mathbb{R}^{I_1}$, a matrix $\boldsymbol{M}\in \mathbb{R}^{I_1\times I_2}$, and a 3$rd$-order tensor $\boldsymbol{\mathcal{T}}\in \mathbb{R}^{I_1\times I_2\times I_3}$. (b) Tensor network representations of matrix trace operation, multiplication and inner product. (c) Tensor network representations of tensor mode-$n$ product, inner product and tensor contraction.}
		\label{img1}
	\end{figure*}
	
	\begin{enumerate}[$\bullet$]
		\item We propose a Multi-Tensor Network Representation (MTNR) framework for high-order tensor decomposition, which can be regarded as a combination of multiple TD models. In theory, we define the fundamental algebraic operations among the tensors with unified TN representation. Note that these properties allow us to perform multilinear operations among tensors represented by different TN models in a latent factor space with linear complexity instead of multilinear space with exponential complexity.
		
		\item We present an adaptive topology learning (ATL) algorithm for MTNR based on a rank incremental strategy and a projection error measurement strategy. The structure of MTNR is data-dependent and can approximate complex low-rank information from different subspaces and obtains a better low-rank representation.
		
		\item We apply MTNR to the typical TC task and develop two efficient algorithms: MTNR-ALS and MTNR-ADMM. The former is basic and based on the Alternating Least Squares (ALS) scheme~\cite{ref55} while the latter is built on the Alternating Direction Method of Multiplier (ADMM) framework~\cite{ref54}, involving latent low-rank regularizers. The effectiveness of MTNR is demonstrated on both synthetic data and real-world datasets via extensive experiments.
	\end{enumerate}

	
	
	
	The remainder of this paper is organized as follows. In Section \ref{sec2}, the preliminaries of tensor notations and operations are introduced. Section \ref{sec3} presents the MTNR framework and TN-based tensor algebraic operations in detail. In Section \ref{sec4}, two algorithms are proposed for solving the LRTC problem. The numerical analysis and experimental comparisons are given in Section \ref{sec5}. Finally, we summarize this work in Section \ref{sec6}.
	
	\section{Notations and Preliminaries}
	\label{sec2}
	
	In this paper, we use boldface calligraphic letters to denote tensors of order $N\geq3$, e.g.,$\ \boldsymbol{\mathcal{A}}$; boldface capital letters to denote matrices, e.g.,$\ \boldsymbol{A }$. The boldface lowercase letters and lowercase letters respectively denote vectors and scalars, e.g.,$\ \boldsymbol{a}$ and $a$. For a N-order tensor $\boldsymbol{\mathcal{A}}\in \mathbb{R} ^{I_1\times \cdots \times I_N}$, its $(i_1,i_2,\cdots,i_N)$-th entry is denote as $\boldsymbol{\mathcal{A}}(i_1,i_2,\cdots,i_N)$ or $\boldsymbol{\mathcal{A}}_{i_1,i_2,\cdots,i_N}$, where $i_n\in [I_n],\ n\in [N]$. The $[N]$ denotes a set including integers from 1 to $N$. The inner product of $\boldsymbol{\mathcal{A}},\boldsymbol{\mathcal{B}}\in \mathbb{R} ^{I_1\times \cdots \times I_N}$ is defined as $\left \langle \boldsymbol{\mathcal{A}}, \boldsymbol{\mathcal{B}}\right\rangle=\sum_{i_1,\cdots,i_N} \boldsymbol{\mathcal {A}}_{i_1,i_2,\cdots,i_N}\boldsymbol{\mathcal{B}}_{i_1,i_2,\cdots,i_N}$. The Frobenius norm and $l_1$ norm of $\boldsymbol{\mathcal{A}}\in \mathbb{R} ^{I_1\times \cdots \times I_N}$ are defined as $||\boldsymbol{\mathcal{A}}||_F=\sqrt{ \left \langle \boldsymbol{\mathcal{A}},\boldsymbol{\mathcal{A}}\right\rangle}$ and $||\boldsymbol{\mathcal{A}}||_1=\sum_{i_1 ,\cdots,i_N} |\boldsymbol{\mathcal{A}}_{i_1,i_2,\cdots,i_N}|$, which also applies to matrices and vectors. For a matrix $\boldsymbol{A}\in \mathbb {R} ^{I_1\times I_2}$, the rank of the $\boldsymbol{A}$ is represented as $rank(\boldsymbol{A})$, and the nuclear norm of $\boldsymbol{A}$ is denoted as $||A||_*=\sum_i \sigma_i(\boldsymbol{ A})$, where $\sigma_i(\boldsymbol{A})$ represents the $i$-th eigenvalue of $ \boldsymbol{A}$. The conjugate transpose, inverse, and pseudo-inverse of matrix $\boldsymbol{{A}}$ are denoted as $\boldsymbol{{A}}^*$, $\boldsymbol{{A}}^{-1}$ and $\boldsymbol{{A}}^\dagger$, respectively. $I_n$ indicates the identity matrix of size $n\times n $. In addition, some definitions of tensor operations \cite{ref1,ref29,ref37} involved in the paper are given as follows.
	\begin{Definition}(Vectorization and Matricization). \it 
		\label{def2.1}
		For an $N$-order tensor $\boldsymbol{\mathcal{A}}\in \mathbb{R} ^{I_1\times \cdots \times I_N}$. The vectorization of $\boldsymbol{\mathcal{A}}$ is denote as $vec(\boldsymbol{\mathcal{A}})\in \mathbb{R}^{\prod I_i}$, with entries $vec(\boldsymbol{\mathcal{A}})_{\overline{i_1\cdots i_N}}=\boldsymbol{\mathcal{A}}(i_1, \cdots, i_N)$. The mode-$k$ matricization (also known as unfolding or flattening) of $\boldsymbol{\mathcal{A}}$ arranges the mode-k fibers into the columns of the resulting matrix, denote as $\boldsymbol{A}_{(k)}\in \mathbb{R}^{I_k\times \prod_{i\neq k} I_i}$ and its elements are given by
		\begin{equation}\begin{split}
				\boldsymbol{A}_{(k)}(i_k, \overline{i_1\cdots i_{k-1}i_{k+1}i_N})=\boldsymbol{\mathcal{A}}(i_1,i_2,\cdots i_N).
				\nonumber
		\end{split}\end{equation}
		Furthermore, the $k$-matricization of $\boldsymbol{\mathcal{A}}$ is denote as $\boldsymbol{{A}}_{\left \langle k \right \rangle }\in \mathbb{R} ^{I_1\cdots I_k\times I_{k+1}\cdots I_N}$, which can be regarded as mode-1 matricization of $\boldsymbol{\mathcal{A}}$ after merging the first $k$ modes and its elements are given by
		\begin{equation}\begin{split}
				\boldsymbol{A}_{\left \langle k \right \rangle }(\overline{i_1,\cdots i_k}, \overline{i_{k+1}\cdots i_N})=\boldsymbol{\mathcal{A}}(i_1,i_2,\cdots i_N).
				\nonumber
		\end{split}\end{equation}
	\end{Definition}
	Note that the multi-indices is defined by little-endian convention~\cite{ref34} as $\overline{i_1i_2\cdots i_N}=1+\sum_{k=1}^N (i_k-1)J_k$ with $J_k=\prod_{m=1}^{k-1}I_m$.
	
	\begin{Definition} (Multilinear Tensor Rank). \it 
		\label{def2.2}
		For an $N$-order tensor $\boldsymbol{\mathcal{A}}\in \mathbb{R} ^{I_1\times \cdots \times I_N}$. The multilinear tensor rank (also known as Tucker rank) of $\boldsymbol{\mathcal{A}}$ is a vector and denoted as $(rank(\boldsymbol{A}_{(1)}), rank(\boldsymbol{A}_{(2)}),\cdots, rank(\boldsymbol{A}_{(N)}))^T$, where $\boldsymbol{A}_{(N)}$ represents mode-$n$ Matricization of $\boldsymbol{\mathcal{A}}$.
	\end{Definition}
	Then the SNN can be defined as $\sum_{i=1}^N ||\boldsymbol{A}_{(i)}||_*$. We further present several basis tensor operations used throughout this paper.
	
	\begin{Definition} (Hadamard Product). \it
		\label{def2.3}
		Let two $N$-order tensors $\boldsymbol{\mathcal{A}}, \boldsymbol{\mathcal{B}}\in \mathbb{R} ^{I_1\times \cdots \times I_N}$. The Hadamard product between these two tensors is element-wise product and defined as $\boldsymbol{\mathcal{C}}=\boldsymbol{\mathcal{A}}\circledast \boldsymbol{\mathcal{B}}\in \mathbb{R} ^{I_1\times \cdots \times I_N}$, with entries $\boldsymbol{\mathcal{C}}_{i_1,i_2,\cdots,i_N}=\boldsymbol{\mathcal{A}}_{i_1,i_2,\cdots,i_N}\boldsymbol{\mathcal{B}}_{i_1,i_2,\cdots,i_N}$.
	\end{Definition}
	
	\begin{Definition} (Kronecker Product). \it 
		\label{def2.4}
		Let two $N$-order tensors $\boldsymbol{\mathcal{A}}\in \mathbb{R} ^{I_1\times \cdots \times I_N}$ and $\boldsymbol{\mathcal{B}}\in \mathbb{R} ^{J_1\times \cdots \times J_N}$. The Kronecker product between these two tensors yields an $N$-order tensor $\boldsymbol{\mathcal{C}}=\boldsymbol{\mathcal{A}}\otimes \boldsymbol{\mathcal{B}}$ of size ${I_1J_1\times \cdots \times I_NJ_N}$, with entries $\boldsymbol{\mathcal{C}}(\overline{i_1j_1},\cdots,\overline{i_Nj_N})=\boldsymbol{\mathcal{A}}(i_1,\cdots i_N)\boldsymbol{\mathcal{B}}(j_1,\cdots j_N).$
	\end{Definition}
	
	\begin{Definition} (Mode-n Khatri-Rao product). \it
		\label{def2.5}
		Let an $N$-order tensor $\boldsymbol{\mathcal{A}}\in \mathbb{R} ^{I_1\times \cdots \times I_N}$ and $\boldsymbol{\mathcal{B}}\in \mathbb{R} ^{J_1\times \cdots \times J_N}$ have $I_n=J_n, n\in [N]$. The mode-n Khatri-Rao product between these two tensors yields an $N$-order tensor $\boldsymbol{\mathcal{C}}=\boldsymbol{\mathcal{A}}\odot_{n} \boldsymbol{\mathcal{B}}$ of size ${I_1J_1\times \cdots \times I_{n-1}J_{n-1}\times I_{n}J_{n}\times I_{n+1}J_{n+1}\times \cdots \times I_{N}J_{N}}$, with entries $\boldsymbol{\mathcal{C}}(\overline{i_{1}j_{1}},\cdots,\overline{i_{n-1}j_{n-1}},i_n,\overline{i_{n+1}j_{n+1}},\cdots,\overline{i_Nj_N})=\boldsymbol{\mathcal{A}}(i_1,\cdots i_N)\boldsymbol{\mathcal{B}}(j_1,\cdots j_N)$ and subtensors $\boldsymbol{\mathcal{C}}(:,\cdots,:,i_n,:,\cdots,:)=\boldsymbol{\mathcal{A}}(:,\cdots,:,i_n,:,\cdots,:)\otimes\boldsymbol{\mathcal{B}}(:,\cdots,:,j_n,:,\cdots,:)$.
	\end{Definition}
	
	\begin{Definition} (Outer Product). \it
		\label{def2.6}
		Let an $N$-order tensor $\boldsymbol{\mathcal{A}}\in \mathbb{R} ^{I_1\times \cdots \times I_N}$ and $M$th-order tensor $\boldsymbol{\mathcal{B}}\in \mathbb{R} ^{J_1\times \cdots \times J_M}$. The outer product between these two tensors yields an $(N+M)$th-order tensor $\boldsymbol{\mathcal{C}}= \boldsymbol{\mathcal{A}}\circ \boldsymbol{\mathcal{B}}$ of size $I_1\times \cdots I_N\times J_1\times \cdots J_M$, with entries $\boldsymbol{\mathcal{X}}_{i_1,\cdots,i_N,j_1,\cdots,j_M}=\boldsymbol{\mathcal{A}}_{i_1,\cdots,i_N}\boldsymbol{\mathcal{B}}_{j_1,\cdots,j_M}$.
	\end{Definition}

	\begin{Definition} (Mode-$n$ product). \it 
		\label{def2.7}
		Let an $N$-order tensor $\boldsymbol{\mathcal{A}}\in \mathbb{R} ^{I_1\times \cdots \times I_N}$ and matrix $\boldsymbol{B}\in \mathbb{R} ^{J\times I_n}$. The mode-$n$ product between these two tensors yields an $N$-order tensor $\boldsymbol{\mathcal{C}}=\boldsymbol{\mathcal{A}}\times_n \boldsymbol{{B}}$ of size $I_1\times \cdots I_{n-1}\times J\times I_{n+1}\times \cdots I_N$, with entries $C_{i_1,\cdots,i_{n-1},j,i_{n+1},\cdots,i_N}=\sum_{i_n=1}^{I_n} \boldsymbol{\mathcal{A}}_{i_1,\cdots,i_{n-1},i_n,i_{n+1},\cdots,i_N}\boldsymbol{\mathcal{B}}_{j,i_n}$.
	\end{Definition}
	Moreover, the complicated relations and operations among tensors can represent graphically and intuitively by TN~\cite{ref34}. As shown in Fig.\ref{img1}, the vertices with $N$ edges denote an $N$th-order tensor, and each edge denotes one mode. The standard edge, connecting two tensors, represents the modes involved in multilinear operations. To summation along the standard edge between two connecting tensors is called tensor contraction.

	\begin{Definition} (Tensor Contraction). \it 
		\label{def2.8}
		For an $N$-order tensor $\boldsymbol{\mathcal{A}}\in \mathbb{R} ^{I_1\times \cdots \times I_N}$ and $M$th-order tensor $\boldsymbol{\mathcal{B}}\in \mathbb{R} ^{J_1\times \cdots \times J_M}$ with $d$ common modes, $I_i=J_i, i\in [d]$. Let $n=(1,\cdots,N)$ and $m=(1,\cdots,M)$. The tensor contraction between these two tensors alone the common modes yields an $(N+M-2d)$th-order tensor $\boldsymbol{\mathcal{C}}=\boldsymbol{\mathcal{A}}\times_{n_1,\cdots, n_d}^{m_1,\cdots, m_d} \boldsymbol{\mathcal{B}}$ of size $I_{d+1}\times \cdots \times I_N\times J_{d+1}\times \cdots \times J_{M}$, with entries $\boldsymbol{\mathcal{C}}_{i_{d+1},\cdots,i_N,j_{d+1},\cdots,j_M}=\sum_{i_1}^{I_1}\cdots\sum_{i_d}^{I_d}\boldsymbol{\mathcal{A}}_{i_1,\cdots,i_d,d_{d+1},\cdots,i_N}\boldsymbol{\mathcal{B}}_{j_1,\cdots,j_d,j_{d+1},\cdots,j_M}$.
	\end{Definition}
	In essence, tensor contraction is a high-dimensional generalization of matrix multiplication~\cite{ref34} since $vec(\boldsymbol{\mathcal{A}}\times_{I_1,\cdots, I_d}^{J_1,\cdots, J_d} \boldsymbol{\mathcal{B}})=vec(\boldsymbol{\mathcal{A}}_{\left \langle d \right \rangle}^T \boldsymbol{\mathcal{B}}_{\left \langle d \right \rangle})$. Thus, the mode-$n$ product and multilinear product can be classified as tensor contraction.
	
	\section{Multi-Tensor Network Representation}
	\label{sec3}
	Equipped with the above preliminaries, we present the MTNR framework in detail in this section. We first introduce the standard TN representation and extend it to multiple cases. Then, an adaptive topology learning algorithm is proposed for MTNR. Finally, the fundamental algebraic operations are established among the tensors with TN representations.
	
	\subsection{Tensor Network Representation}
	The TN model represents a high-order tensor as the multilinear operation over a set of latent factors, and the number of factors is equivalent to the order of the tensor without considering the internal vertices. As illustrated in Fig.\ref{img2}, the difference among various TD models lies in the topological structure constructed by the connections among latent factors, e.g., the TR model~\cite{ref37} establishes a connection between the head and tail factors based on TT~\cite{ref36}. For the sake of clarity, we employ $\Psi$ to denote the TN model uniformly. Suppose that an $N$-order tensor $\boldsymbol{\mathcal{X}}\in \mathbb{R} ^{I_1\times \cdots \times I_N}$ is decomposed by $\Psi$ and we obtain $N$ factors $\boldsymbol{\mathcal{Z}}^{(k )}\in \mathbb{R}^{R_{1,k}\times \dots \times R_{k-1, k}\times I_k \times R_{k+1, k}\times \dots\times R_{ N,k}}$ with $k\in [N]$. Hence, the rank of $\Psi$ can be naturally expressed as a matrix $\boldsymbol{R}_\Psi \in\mathbb{R}^{(N-1)\times (N-1)}$, which is
	\begin{equation}\begin{split}
			\boldsymbol{R}_\Psi =
			\begin{pmatrix}
				R_{1,2} &  R_{1,3}&  \cdots & R_{1,N}\\
				R_{2,1} &  R_{2,3}& \cdots &  R_{2,N} \\
				\vdots  &  \vdots &  \ddots & \vdots\\
				R_{N-1,1} &R_{N-2,2}  & \cdots & R_{N-1,N}
			\end{pmatrix},
			\label{eq4}
	\end{split}\end{equation}
	where $R_{i,j}$ represents the dimension of common mode between $\boldsymbol{\mathcal{Z}}^{(i)}$ and $\boldsymbol{\mathcal{Z}}^{(j)}$. We call it edge rank and it satisfies $R_{i,j}=R_{j,i}$($1\leq i< j \leq N, i,j\in \mathbb{N}_+$). The element-wise form of $\Psi$ decomposition can be expressed as
	\begin{equation}\begin{split}
			\boldsymbol{\mathcal{X}}&(i_1,i_2,...,i_N)=\Re(\boldsymbol{\mathcal{Z}}^{(1)}, \cdots, \boldsymbol{\mathcal{Z}}^{(N)})(i_1,i_2,...,i_N)\\
			&=\sum_{r_{1,2}=1}^{R_{1,2}}\cdots\sum_{r_{1,N}=1}^{R_{1,N}}\sum_{r_{2,3}=1}^{R_{2,3}}\cdots\sum_{r_{2,N}=1}^{R_{2,N}}\cdots\sum_{r_{N-1,N}=1}^{R_{N-1,N}}\\
			&\boldsymbol{\mathcal{Z}}^{(1)}_{i_1,r_{1,2},...,r_{1,N}}\boldsymbol{\mathcal{Z}}^{(2)}_{r_{2,1},i_2,...,r_{2,N}}\cdots\boldsymbol{\mathcal{Z}}^{(N)}_{r_{N,1},...,r_{N,N-1},i_N},
			\label{eq5}
	\end{split}\end{equation}
	where $\Re$ denotes tensor contraction operation and $\Re(\boldsymbol{\mathcal{Z}}^{(1)}, \cdots, \boldsymbol{\mathcal{Z}}^{(N)})$ indicates the recovered tensor generated by factors. Then we introduce the new definition as follows:
	
	\begin{Definition} (Subnetwork). \it 
		\label{def3.1}
		Let an $N$-order tensor $\boldsymbol{\mathcal{A}}\in \mathbb{R} ^{I_1\times \cdots \times I_N}$ be decomposed by $\Psi$ and has TN representations $\boldsymbol{\mathcal{A}}=\Re(\boldsymbol{\mathcal{Z}}^{(1)},\cdots, \boldsymbol{\mathcal{Z}}^{(N)} )$, where $\boldsymbol{\mathcal{Z}}^{(k)}\in \mathbb{R} ^{R_{1,k}\times \dots \times I_k \times \dots\times R_{N,k}}$ for $k=1,\cdots, N$. Given a vector $\boldsymbol{n} \in \mathbb{R}^{N}$, which is reordering of $(1,\cdots, N)$, then the contraction of subnetwork that consisting of the factors $\{\boldsymbol{\mathcal{Z}}^{(n_1)},\cdots, \boldsymbol{\mathcal{Z}}^{(n_r)}  \},1\leq r\leq N,$ will obtain an $N$-order tensor $\Re(\boldsymbol{\mathcal{Z}}^{(n_1)},\cdots, \boldsymbol{\mathcal{Z}}^{(n_r)} )$ of size $I_{n_1}\times\cdots\times I_{n_r}\times R_{n_1,n_{r+1}}\cdots R_{n_r,n_{r+1}}\times R_{n_1,n_{r+2}}\cdots R_{n_r,n_{r+2}}\times\cdots\times R_{n_1,n_{N}}\cdots R_{n_r,n_{N}}$.
	\end{Definition}

	The calculation of $\Re$ in (\ref{eq5}) can be completed with $N-1$ times matrix multiplication because the mode-n matricization of $\boldsymbol{\mathcal{X}}$ can be written as
	\begin{equation}\begin{split}
			\boldsymbol{{X}}_{(n)}= \boldsymbol{{Z}}^{(n)}_{(n)} (\boldsymbol{{Z}}^{(\neq n)}_{\left \langle N-1 \right \rangle })^T,
			\label{eq6}
	\end{split}\end{equation}
	where $\boldsymbol{\mathcal{Z}}^{(\neq n)}=\Re(\{\boldsymbol{\mathcal{Z}}^{(k)} \}_{ ^{k=1}_{k\neq n}}^N)$
	is an $N$-order tensor of size $I_1\times\cdots\times I_{n-1}\times I_{n+1}\times\cdots\times I_{N}\times R_{1,n}\cdots R_{n-1,n} R_{n,n+1}\cdots R_{n,N}$, obtained by contracting the subnetwork of $\Psi$ composed of N-1 factors except $\boldsymbol{\mathcal{Z}}^{(n)}$. Similarly, the $\boldsymbol{\mathcal{Z}}^{(\neq \{m,n\})}, \boldsymbol{\mathcal{Z}}^{(\leq m)}$, and $\boldsymbol{\mathcal{Z}}^{(>m)}$ are defined as $\Re(\{\boldsymbol{\mathcal{Z}}^{(k)} \}_{ ^{k=1}_{k\neq m,n}}^N), \Re(\{\boldsymbol{\mathcal{Z}}^{(k)} \}_{{k=1}}^m)$, and $ \Re(\{\boldsymbol{\mathcal{Z}}^{(k)} \}_{ {k=m+1}}^N)$, respectively. 
	
	Note that the TN representation (\ref{eq5}) does not demand a concrete topology strictly, and $\Psi$ conceptually generalizes a family of TD models, including TT and TR representation. This unified representation gives us the chance to learn adaptive decomposition topologies for various data. Another tensor form of $\Psi$ decomposition can be formulated as
	\begin{equation}\begin{split}
			\boldsymbol{\mathcal{X}}&=\sum_{r_{1,2}=1}^{R_{1,2}}\cdots\sum_{r_{1,N}=1}^{R_{1,N}}\sum_{r_{2,3}=1}^{R_{2,3}}\cdots\sum_{r_{2,N}=1}^{R_{2,N}}\cdots\sum_{r_{N-1,N}=1}^{R_{N-1,N}}\\
			& \boldsymbol{z}^{(1)}(r_{1,2},r_{1,3},\cdots,r_{1,N})\circ \boldsymbol{z}^{(2)}(r_{1,2},r_{2,3},\cdots,r_{2,N})\\
			&\circ\cdots\circ\boldsymbol{z}^{(N)}(r_{1,N},r_{2,N},\cdots,r_{N,N-1}),
			\label{eq7}
	\end{split}\end{equation}
	where $\boldsymbol{z}^{(k)}(r_{1,k},\cdots,r_{k-1,k},r_{k,k+1},\cdots,r_{k,N})\in \mathbb{R}^{I_k}$ denote $\overline{r_{1,k}\cdots r_{k-1,k}r_{k,k+1} \cdots r_{k,N}}$th mode-$k$ fiber of factor $\boldsymbol{\mathcal{Z}}^{(k)}$. Then the reconstructed tensors $\boldsymbol{\mathcal{X}} $ can be expressed as the superposition of the rank-1 tensors. In general, TN representation with arbitrary topology is essentially an undirected graph composed of several vertices and edges, and exist a bijective with the simple graph~\cite{ref49}. If condition $R_{i,j}>1, 1\leq i< j\leq N$ is satisfied, then the TN representation of $\Psi$ can be regard as a complete graph, as illustrated graphically in Fig.\ref{img2}(f). 
	\begin{Remark}\it
		All the rank-1 edge is removed before the tensor network contraction since those edges are meanless and do not affect the computation consistency. Hence, two connected factors indicate the corresponding edge rank is larger than one, and the isolated subnetwork is calculated by the outer product~\cite{ref49}. In addition, the self-loops and multiple edges structure is prohibited to avoid confusion.
	\end{Remark}
	\vspace{-1mm}
	Since the topology of the $\Psi$ is not strictly limited. For the TN representation with $N$ factors, there exist $2^{\frac{N(N-1)}{2}}$ kind of topology when taking no account of the edge rank magnitude, from the $null\ graph$ (e.g., TN representation with $N$ isolated vertices) to a fully connected $complete\ graph$. Hence, the selection of TD models can also be viewed as a topology searching process, which is a discrete optimization problem analogous to the layer-wise architecture search in convolutional neural networks~\cite{ref50}. From (\ref{eq5}), we can observe that the edge rank determines the topology and complexity of $\Psi$. Supposing that all the elements in $R_{\Psi}$ are equal to $R$ and $I_1=\cdots=I_N=I$, then the number of parameters requires of $\Psi$ decomposition is $\mathcal{O}(NIR^{N-1})$ and the computational complexity of operation $\Re$ is $\mathcal{O}(\sum_{k=2}^N I^kR^{k(N-k)+k-1})$. However, such storage complexity increases exponentially with the tensor order $N$ and the contraction is computationally too expensive for large-scale data. In order to solve this problem, we set the maximum connections of each latent factor as $t$, where $t\in \mathbb{N}^+$ is a constant and satisfies $1\leq t << N$. The existence of $t$ decreases the degree of freedom of feasible topology and contraction complexity. Thus, the $\Psi$ decomposition required parameters are $\mathcal{O}(NIR^t)$, which is linear with the tensor order, and this property reduces the complexity of calculation and storage significantly.
	
	The relationship of the rank of $\Psi$ and multilinear rank is established in the following theorem.
	\vspace{-1mm}
	\begin{Theorem}\it Let an $N$-order tensor $\boldsymbol{\mathcal{X}}\in \mathbb{R} ^{I_1\times \cdots \times I_N}$ be represented by Equation (\ref{eq5}), then the following two inequalities hold for $n=1,\cdots,N$:
		\begin{equation}\begin{split}
				& rank(\boldsymbol{X}_{(n)}) \leq \prod_{i=1,i\neq n}^{N} R_{n,i},\\
				& rank(\boldsymbol{X}_{\left \langle n \right \rangle})\leq \prod_{i=1}^n\prod_{j=n+1}^N R_{i,j}.  
				\label{eq8}
		\end{split}\end{equation}
	\end{Theorem}
	As a side note, the generalized TN representation defined in this section prohibits the existence of internal factors (e.g., the cores tensor in Tucker decomposition) since they can be eliminated by tensor renormalization group (TRG)~\cite{ref57} algorithm or tensor contraction strategies. For instance, an efficient TR decomposition algorithm is proposed in~\cite{ref56} base on Tucker compression.
	
	\begin{figure}[h]
		\centering
			\includegraphics[width=.5\textwidth]{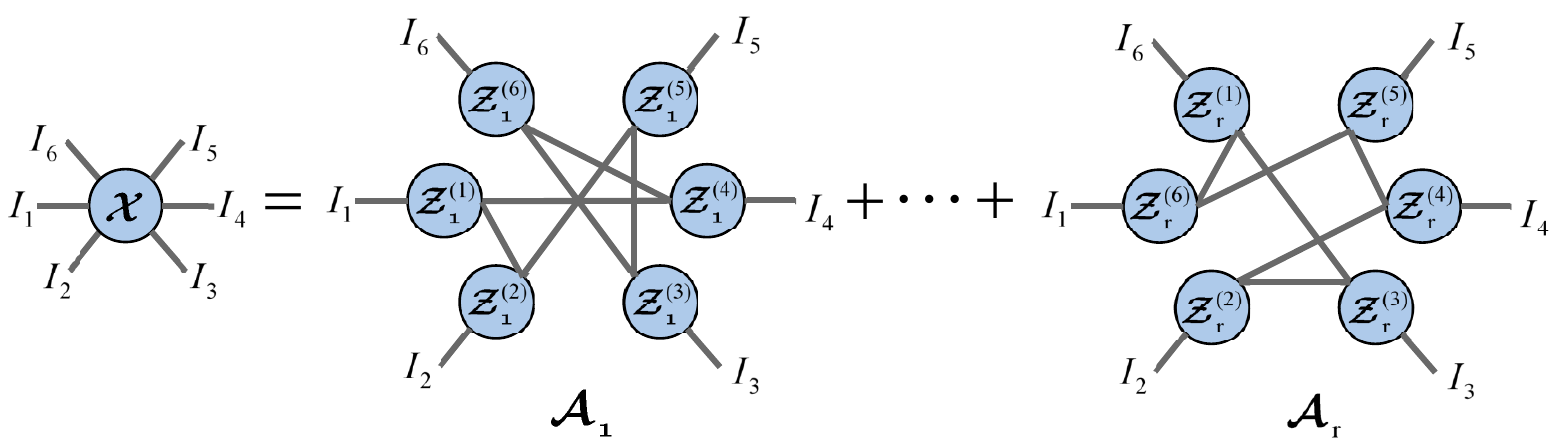}
		\caption{A graphical instance of the MTNR. The topology of each low-rank component's TN representation is irregular, and the maximum number of connections for the factor is set as 2.}
		\label{img3}
	\end{figure}
	
	\subsection{Multi-Tensor Network Representation}
	Based on subspace clustering~\cite{ref53} theory, high-dimensional data can be regarded as the combination of low-rank components from different subspaces. As such, an $N$-order tensors $\boldsymbol{\mathcal{X}}\in \mathbb{R} ^{I_1 \times \cdots \times I_N}$ can be represented by an addition of multiple components as
	\begin{equation}\begin{split}
			\boldsymbol{\mathcal{X}}=\boldsymbol{\mathcal{A}}_1+\boldsymbol{\mathcal{A}}_2+\cdots+\boldsymbol{\mathcal{A}}_r,
			\label{eq9}
	\end{split}\end{equation}
	where $\boldsymbol{\mathcal{A}}_i \in \mathbb{R} ^{I_1 \times \cdots \times I_N}$ represents the $i$th low-rank component for $i=1,\cdots,r$. We can observe that if each component $A_i$ is a rank-1 tensor, then (\ref{eq9}) is equivalent to CP decomposition and $r$ is just CP-rank.
	
	Most of the high-dimensional data are complicated and mode-interrelated. The TD model with simple topology (e.g., ring-shaped TR) to capture the intrinsic low-rank structure inside the data and only require linear storage complexity~\cite{ref37}. However, the connections involved in topology only exist in the adjacent factors, leading to exponential correlation decay and expecting a larger rank to capture more global features. On the contrary, the TD model with complex topology can achieve better results, e.g., FCTN~\cite{ref39}, but the storage and computational complexity are exponential with the order of the data. In order to better weigh the representation capability and complexity of TN models. As shown in Fig.\ref{img3}, MTNR aims to discover TN topologies from data and manipulate multiple TD models with different topologies to decompose components separately, which is critical compared with other TD models. Then, (\ref{eq9}) can be rewritten as
	\begin{equation}\begin{split}
			\boldsymbol{\mathcal{X}}=\Re(\Psi_1(\boldsymbol{\mathcal{A}}_1))+\Re(\Psi_2(\boldsymbol{\mathcal{A}}_2))+\cdots+\Re(\Psi_r(\boldsymbol{\mathcal{A}}_r)),
			\label{eq10}
	\end{split}\end{equation}
	where $\Psi_i$ represents the $i$th TD models with a simple topological structure for $i=1,\cdots,r$, and the corresponding latent factor $\{\boldsymbol{\mathcal{Z} }^{(k)}_i \}_{k=1}^N$ is obtained via $\Psi_i(\boldsymbol{\mathcal{A}}_i)$. From (\ref{eq10}), we see that MTNR can be considered as a combination of multiple TD models, which provides MTNR with more compelling representation and generalization capabilities. The amount of parameters for MTNR is $\mathcal{O}(rNIR^t)$, which is linear to the tensor order. Note that the structure of each TN model in MTNR is data-adaptive to better capture the low-rank structure and reduce approximation deviation. Besides, the combination of multiple identical TD models retains the properties of a single one, such as inverse permutation invariance in TT, and circular dimensional permutation invariance in TR.

	\subsection{Adaptive Topology Learning Algorithm}
	A key problem of low-rank approximation based on TN representation is to search adaptive topology for various complicated and mode-imbalanced data. Existing data-driven approaches are difficult for data with different orders and sizes (such as neural architecture search and reinforcement learning). In this part, we present a novel efficient adaptive topology learning (ATL) algorithm for MTNR.
	
	The main objective of ATL is to obtain a low-rank approximation of data with less storage resources. Given an $N$-order tensor $\boldsymbol{\mathcal{X}}\in \mathbb{R} ^{I_1\times \cdots \times I_N}$ and relative error $\epsilon$, the optimization problems of MTNR can be formulated as
	\begin{equation}\begin{split}
			&\min_{\boldsymbol{\mathcal{Z}}^{(1)}_1,\cdots,\boldsymbol{\mathcal{Z}}^{(N)}_r}\quad \mathcal{P}(\boldsymbol{\mathcal{Z}}^{(1)}_1,\cdots,\boldsymbol{\mathcal{Z}}^{(N)}_r)\\
			\text{s.t.}\ \  &||\boldsymbol{\mathcal{X}} - \sum_{i=1}^r \Re(\boldsymbol{\mathcal{Z}}^{(1)}_i,\cdots, \boldsymbol{\mathcal{Z}}^{(N)}_i)||_F\leq \epsilon ||\boldsymbol{\mathcal{X}}||_F\\
			&\ \  \mathcal{P}(\boldsymbol{\mathcal{Z}}^{(1)}_i,\cdots,\boldsymbol{\mathcal{Z}}^{(N)}_i)\leq \gamma_i,\ i=1,\cdots,r.
			\label{eq11}
	\end{split}\end{equation}
	Here $\mathcal{P}(\cdot)$ denotes storage cost function and $\gamma_i$ is the predefined upper bound of the $i$th low-rank component representation. In fact, instead of specifying the number of components $r$, we choose to sequentially optimize each low-rank component to fit the current residual estimation, which can reduce the complexity of the topological search and obtain more stable outcomes. To solve the optimization problem of each low-rank component is based on the following two strategies.\par
	\vspace{2mm}

	\textbf{1) Rank incremental strategy}.
	The rank incremental strategy~\cite{ref47, ref48} suggests that tensors with TN representation starts from the rank-1 tensor, successively selects a factor pair and increases its corresponding edge rank to greatly reduce approximate error. The topology and ranks obtained in this manner have better data adaptability. Specifically, for the $k$th low-rank component, $\boldsymbol{\mathcal{A}}_k=\Re(\boldsymbol{\mathcal{Z}}^{(1)}_k,\cdots, \boldsymbol{\mathcal{Z}}^{(N)}_k)$, the corresponding optimization problem can be derived from (\ref{eq11}) as
	\begin{equation}\begin{split}
			\min_{\boldsymbol{\mathcal{Z}}^{(1)}_k,\cdots,\boldsymbol{\mathcal{Z}}^{(N)}_k}&\quad ||\boldsymbol{\mathcal{T}}- \Re(\boldsymbol{\mathcal{Z}}^{(1)}_k,\cdots, \boldsymbol{\mathcal{Z}}^{(N)}_k)||_F^2\\
			\text{s.t.}\ \  &\mathcal{P}(\boldsymbol{\mathcal{Z}}^{(1)}_k,\cdots,\boldsymbol{\mathcal{Z}}^{(N)}_k)\leq \gamma_k,
			\label{eq12}
	\end{split}\end{equation}
	where $\boldsymbol{\mathcal{T}}=\boldsymbol{\mathcal{X}}-\sum_{i=1}^{k-1} \Re(\boldsymbol{\mathcal{Z}}^{(1)}_i,\cdots, \boldsymbol{\mathcal{Z}}^{(N)}_i)$ represents residual estimation tensor. The (\ref{eq12}) includes $N$ inter-coupled variables, so the well-known ALS~\cite{ref37} scheme or gradient descent algorithm~\cite{ref58} can be exploited to obtain the numerical solution. The principle of the ALS scheme is to update each variable sequentially while fixing the others until all the variables satisfy the convergence criterion. Then the subproblem w.r.t.the $n$th factor variable $\boldsymbol{\mathcal{Z}}^{(n)}_k\in \mathbb{R}^{R_{n,1}^k\times \dots R_{n ,n-1}^k\times I_n \times R_{n,n+1}^k \dots\times R_{n,N}^k}$ can be expressed in a mode-$n$ matrix form as
	\begin{equation}\begin{split}
			\min_{\boldsymbol{\mathcal{Z}}^{(n)}_k} \quad || \boldsymbol{{T}}_{(n)} - \boldsymbol{{Z}}^{(n)}_{k(n)} (\boldsymbol{{Z}}^{(\neq n)}_{k\left \langle N-1 \right \rangle })^T ||_F^2,
			\label{eq13}
	\end{split}\end{equation}
	which is a least-square optimization problem and has a closed-form solution
	\begin{equation}\begin{split}
			\boldsymbol{{Z}}^{(n)}_{k(n)} = \boldsymbol{{T}}_{(n)} \boldsymbol{{Z}}^{(\neq n)}_{k\left \langle N-1 \right \rangle }  ((\boldsymbol{{Z}}^{(\neq n)}_{k\left \langle N-1 \right \rangle })^T\boldsymbol{{Z}}^{(\neq n)}_{k\left \langle N-1 \right \rangle })^\dagger.
			\label{eq14}
	\end{split}\end{equation}
	All factors $\{\boldsymbol{\mathcal{Z}}^{(1)}_k,\cdots, \boldsymbol{\mathcal{Z}}^{(N)}_k \}$ are circulantly updated until satisfy $\frac{||\boldsymbol{\mathcal{A}}_k^{S+1}-\boldsymbol{\mathcal{A}}_k^{S}||_F}{\boldsymbol{\mathcal{A}}_k^{S}}\leq \delta $, where $S$ is current iteration number and $\delta$ is a specific error level. In this context, we will identify a promising factor pair and increase its corresponding edge rank according to the following projection error measurement strategy if the complexity restraints are satisfied.
	
	\vspace{2mm}
	\textbf{2) Projection error measurement strategy}. The residual tensor $\boldsymbol{\mathcal{T}}$ of size $I_1\times \cdots \times I_N$ is approximated by the $k$th low-rank component $\boldsymbol{\mathcal{A}}_k $, with TN representation $\Re(\boldsymbol{\mathcal{Z}}^{(1)}_k,\cdots, \boldsymbol{\mathcal{Z}}^{(N)}_k) $, where $\boldsymbol{\mathcal{Z}}^{(i)}_k\in \mathbb{R} ^{R_{i,1}^k\times \dots R_{i,i-1}^k\times I_i \times \dots\times R_{i,N}^k}$. Since the correlation within different modes of the reconstructed tensor is established through the connection within the factors, we can increase the edge rank $R_{i,j}^k$ to improve the correlation between the $i$ and $j$ modes of $\boldsymbol{\mathcal{A}}_k$. Each selection for an $N$-order tensor produces $\frac{N(N-1)}{2}$ cases. A simple technique is greedy search~\cite{ref47}, which uses brute-force search to test all options and for each trial it requires several additional iterations to return the most promising edges to improve the model performance. However, that is too computationally expensive. Inspired by~\cite{ref48}, we propose an improved projection error measurement strategy for identifying the optimal factor pair. For $i,j\in \mathbb{N}^+$ satisfies $1\leq i< j\leq N$, consider the following problem
	\begin{equation}\begin{split}
			\min_{W_{i,j}}\quad ||\boldsymbol{E}_{(i,j)}  - \boldsymbol{Z}^{(i,j)}_{k \left \langle 2\right \rangle } W_{i,j}||_F^2,
			\label{eq15}
	\end{split}\end{equation}
	where $\boldsymbol{E}_{(i,j)}$ of size ${I_i I_j\times \prod_{k\neq i,j}I_k}$ denotes the mode-$(i,j)$ matricization of the approximation error tensor $\boldsymbol{\mathcal{T}} -\boldsymbol{\mathcal{A}}_k$ , and $W_{i,j}$ is viewed as the error projection matrix on the factors $\boldsymbol{Z}^{(i)}_{k}$ and $\boldsymbol{Z}^{(j)}_{k}$, computed by $W_{i,j}=(\boldsymbol{Z}^{(\{i,j\})}_{k \left \langle 2 \right \rangle })^\dagger \boldsymbol{E}_{\left \langle i,j \right \rangle }$. We use the $w_{i,j}={||W_{i,j}||_F}$, which we called projection error measurement, to evaluate the importance or sensitivity of the factors pair $(\boldsymbol{Z}^{(i)}_{k},\boldsymbol{Z}^{(j)}_{k})$. In each iteration, the order of factors is updated randomly to evade the influence of mode arrangement. In section~\ref{sec5}, we demonstrate $w$ can effectively measure the importance of the local topology of the TN representation. Moreover, we set three conditions to guarantee the effectiveness of the selection process. 
	
	\noindent(\romannumeral1) Refuse to select the identical factor pair twice consecutively. 
	
	\noindent(\romannumeral2) The number of selections for each factor pair is less than the corresponding mode dimension.
	
	\noindent(\romannumeral3) The maximal connections of each factor are set as $t$. 
	
	The above conditions are used to ensure the diversity and feasibility of the TD topologies generated by ATL. In addition, these topologies are robust to the arrangement of data modes. Overall, we will continuously collect the low-rank components until the stop criterion $||\boldsymbol{\mathcal{X}}-\sum_{i=1}^r \boldsymbol{\mathcal{A}}_i||_F\leq \epsilon||\boldsymbol{\mathcal{X}}||_F$ holds. Algorithm~\ref{alg1} summarizes the whole learning process of MTNR in detail. The significance of the low-rank components obtained by MTNR is decreasing gradually, and the corresponding TN topology is expected to seek the optimal one under the current structure error. In this way, MTNR is prone to form multiple TN models with different topologies with linear parameters.
	
	\floatname{algorithm}{Algorithm}
	\renewcommand{\algorithmicrequire}{\textbf{Input:}}  
	\renewcommand{\algorithmicensure}{\textbf{Output:}}    
	\begin{algorithm}[htbp]
		\caption{: ATL Algorithm for MTNR}
		\begin{algorithmic}[1]
			\label{alg1}
			\REQUIRE $N$-order tensor $\boldsymbol{\mathcal{X}}\in \mathbb{R} ^{I_1\times \dots \times I_N}$, relative error $\epsilon$, complexity restrain $\gamma,  S_{max}$, and threshold $\delta$.
			\ENSURE Factors $\{\boldsymbol{\mathcal{Z}}^{(1)}_1,\cdots,\boldsymbol{\mathcal{Z}}^{(N)}_1,\boldsymbol{\mathcal{Z}}^{(1)}_2,\cdots,\boldsymbol{\mathcal{Z}}^{(N)}_2,\cdots \}$.
			\STATE $k\gets 1$, $\boldsymbol{\mathcal{T}}=\boldsymbol{\mathcal{X}}$.
			\REPEAT
			\STATE Initialize $R_{(i,j)}^k\gets 1 $ for $ 1\leq i < j \leq N$, $\boldsymbol{\mathcal{Z}}_k^{(n)} \in \mathbb{R} ^{R_{(1,n)}^k\times\cdots R_{(n-1,n)}^k \times I_n\times R_{(n,n+1)}^k \dots \times R_{(n,N)}^k}$ for $n=1,\cdots,N$.
			\FOR{$s$=1 to $S_{max}$}
			\STATE $\boldsymbol{\mathcal{A}}_k\gets \Re(\boldsymbol{\mathcal{Z}}_k^{(1)},\cdots, \boldsymbol{\mathcal{Z}}_k^{(N)})$.
			\STATE Update $\boldsymbol{\mathcal{Z}}_k^{(n)}$ via (\ref{eq14}), $n=1,\cdots, N$.
			\STATE $rse\gets \frac{||\Re(\boldsymbol{\mathcal{Z}}_k^{(1)},\cdots, \boldsymbol{\mathcal{Z}}_k^{(N)})  - \boldsymbol{\mathcal{A}}_k ||_F}{||\boldsymbol{\mathcal{A}}_k||_F}$
			\IF{$rse \leq \delta$ and ${\mathcal{P} }(\boldsymbol{\mathcal{Z}}_k^{(1)},\cdots, \boldsymbol{\mathcal{Z}}_k^{(N)})\leq\gamma$}
			\STATE Select factor pair $(\boldsymbol{\mathcal{Z}}_k^{(i)}, \boldsymbol{\mathcal{Z}}_k^{(j)}) $ via (\ref{eq15}).
			\STATE $R_{(i,j)}^k\gets R_{(i,j)}^k+1$.
			\STATE Add a new slice to $j$th mode of $\boldsymbol{\mathcal{Z}}_k^{(i)}$.
			\STATE Add a new slice to $i$th mode of $\boldsymbol{\mathcal{Z}}_k^{(j)}$.
			\ENDIF 
			\ENDFOR
			\STATE $\boldsymbol{{\mathcal{T}}}\gets \boldsymbol{{\mathcal{T}}} - \boldsymbol{\mathcal{A}}_k, k\gets k+1$.
			\UNTIL The convergence condition $\frac{|| \boldsymbol{\mathcal{T}} ||_F}{|| \boldsymbol{\mathcal{A}}_k||_F} \leq \epsilon$ is achieved.
		\end{algorithmic}
	\end{algorithm}
	
	\vspace{-4mm}
	\subsection{Multilinear operations with TN representation}
	An interesting conjecture is whether it is possible to perform multilinear operations in a latent factor space instead of a multilinear space for tensors with TN representation. This section gives the affirmative answer and theoretically establishes basic algebra operations, including multilinear, Hadamard, outer, inner, Kronecker, and mode-n Khatri-Rao products and addition operation. These properties allow us to process large-scale data with linear storage cost and significantly reduce the computational complexity. In~\cite{ref36,ref37}, the authors establish basic multilinear operations for the tensors with TT and TR format, and here we further extend it to a generalized scenario and thus we have the following propositions.
	
	\begin{Proposition} 
		\label{th2}
		\it Suppose that an $N$-order tensor $\boldsymbol{\mathcal{A}}\in \mathbb{R} ^{I_1\times \cdots \times I_N}$ is decomposed by $\Psi$, having a TN representation of $\boldsymbol{\mathcal{A}}=\Re(\boldsymbol{\mathcal{Z}}^{(1)},\cdots, \boldsymbol{\mathcal{Z}}^{(N)} )$, where $\boldsymbol{\mathcal{Z}}^{(k)}\in \mathbb{R}^{R_{k,1}\times \cdots \times R_{k,k-1}\times I_k\times R_{k,k+1}\times \cdots \times R_{k,N}}$. Then the mode-$n$ product (see Def.~\ref{def2.7}) of $\boldsymbol{\mathcal{A}}$ and a matrix $\boldsymbol{B}\in \mathbb{R} ^{J\times I_n}$ can be written as:
		\begin{equation}\begin{split}
				\boldsymbol{\mathcal{A}}\times_n \boldsymbol{B}&=\Re(\boldsymbol{\mathcal{Z}}^{(1)},\cdots, \boldsymbol{\mathcal{Z}}^{(N)} )\times_n \boldsymbol{B}\\
				&=\Re(\Re(\{\boldsymbol{\mathcal{Z}}^{(k)} \}_{ ^{k=1}_{k\neq n}}^N),\boldsymbol{\mathcal{Z}}^{(n)}\times_n \boldsymbol{B})\\
				&=\Re(\boldsymbol{\mathcal{Z}}^{(1)}, \cdots, \boldsymbol{\mathcal{Z}}^{(n)}\times_n \boldsymbol{B},\cdots,\boldsymbol{\mathcal{Z}}^{(N)}).
				\label{eq16}
		\end{split}\end{equation}
		Moreover, the multilinear product of $\boldsymbol{\mathcal{A}}$ with a set of vectors, $u_k\in \mathbb{R}^{I_n}$ for $k\in[N]$, can be computed by
		\begin{equation}\begin{split}
				\boldsymbol{\mathcal{A}}&\times_1 u_1^T\times_2\cdots \times_N u_N^T\\
				&=\Re(\boldsymbol{\mathcal{Z}}^{(1)},\cdots, \boldsymbol{\mathcal{Z}}^{(N)} )\times_1 u_1^T\times_2\cdots \times_N u_N^T\\
				&=\Re(\boldsymbol{\mathcal{Z}}^{(1)}\times_1 u_1^T, \cdots, \boldsymbol{\mathcal{Z}}^{(N)}\times_N u_N^T).
				\label{eq17}
		\end{split}\end{equation}
	\end{Proposition}
	
	The multilinear product is performed on each factor and its result is also given in a TN format. For instance, if the number of connections of each factor is no more than $t$, then the computational complexity of (\ref{eq16}) and (\ref{eq17}) is equivalent to $\mathcal{O}(IR^tJ)$ and $\mathcal{O}(NIR^t)$, respectively.
	
	\begin{Proposition}\it
		Let two $N$-order tensors $\boldsymbol{\mathcal{A}}, \boldsymbol{\mathcal{B}}\in \mathbb{R} ^{I_1\times \cdots \times I_N}$ be respectively decomposed by $\Psi_1$ and $\Psi_2$, which have TN representations $\boldsymbol{\mathcal{A}}=\Re(\boldsymbol{\mathcal{Z}}^{(1)},\cdots, \boldsymbol{\mathcal{Z}}^{(N)} )$ and $\boldsymbol{\mathcal{B}}=\Re(\boldsymbol{\mathcal{G}}^{(1)},\cdots, \boldsymbol{\mathcal{G}}^{(N)} )$, where $\boldsymbol{\mathcal{Z}}^{(k)}\in \mathbb{R} ^{R_{1,k}\times \dots \times I_k \times \dots\times R_{N,k}}$ and $\boldsymbol{\mathcal{G}}^{(k)}\in \mathbb{R} ^{S_{1,k}\times \dots \times I_k \times \dots\times S_{N,k}}$ for $k=1,\cdots, N$. 
		
		\textbf{1)} The Hadamard product (see Def.~\ref{def2.3}) of $\boldsymbol{\mathcal{A}}$ and $\boldsymbol{\mathcal{B}}$ yields a tensor $\boldsymbol{\mathcal{C}}=\boldsymbol{\mathcal{A}}\circledast \boldsymbol{\mathcal{B}}$, having a TN representation of $\boldsymbol{\mathcal{C}}=\Re(\boldsymbol{\mathcal{Y}}^{(1)},\cdots, \boldsymbol{\mathcal{Y}}^{(N)})$, where each factor $\boldsymbol{\mathcal{Y}}^{(k)}$ of size $R_{1,k}S_{1,k}\times \dots \times I_k \times \dots\times R_{N,k}S_{N,k} $, for $k=1,\cdots,N$, can be computed by 
		\begin{equation}\begin{split}
				\boldsymbol{\mathcal{Y}}^{(k)}=\boldsymbol{\mathcal{Z}}^{(k)}\odot_k \boldsymbol{\mathcal{G}}^{(k)}.
				\label{eq18}
		\end{split}\end{equation}
		where $\boldsymbol{\mathcal{Z}}^{(k)}(i_d)\in \mathbb{R} ^{R_{1,k}\times \dots \times 1 \times \dots\times R_{N,k}}$ is $i_d$th subtensor of $\boldsymbol{\mathcal{Z}}^{(k)}$ in the $k$th mode.
		\rm
		
		\noindent Proof. The n-matricization of $\boldsymbol{\mathcal{A}}$ and $ \boldsymbol{\mathcal{B}}$ can be formulated as $\boldsymbol{{Z}}^{(\leq n)}_{\left \langle n \right \rangle } (\boldsymbol{{Z}}^{(> n)}_{\left \langle N-n \right \rangle })^T $ and $\boldsymbol{{g}}^{(\leq n)}_{\left \langle n \right \rangle } (\boldsymbol{{g}}^{(> n)}_{\left \langle N-n \right \rangle })^T $, then the following equation holds
		\begin{equation}\begin{split}
				(\boldsymbol{\mathcal{A}}\circledast \boldsymbol{\mathcal{B}})_{\left \langle n \right \rangle}=&
				\left ( \boldsymbol{{Z}}^{(\leq n)}_{\left \langle n \right \rangle } (\boldsymbol{{Z}}^{(> n)}_{\left \langle N-n \right \rangle })^T \right ) \circledast  \left (\boldsymbol{{g}}^{(\leq n)}_{\left \langle n \right \rangle } (\boldsymbol{{g}}^{(> n)}_{\left \langle N-n \right \rangle })^T  \right ) \\
				=&\left (\Re(\{\boldsymbol{\mathcal{Z}}^{(k)} \}_{ {k=1}}^n )_{\left \langle n \right \rangle} (\Re( \{\boldsymbol{\mathcal{Z}}^{(k)} \}_{ {k=n+1}}^N )_{\left \langle N-n \right \rangle})^T\right ) \circledast\\
				&\left (\Re(\{\boldsymbol{\mathcal{G}}^{(k)} \}_{ {k=1}}^n )_{\left \langle n \right \rangle} (\Re(\{ \boldsymbol{\mathcal{G}}^{(k)} \}_{ {k=n+1}}^N )_{\left \langle N-n \right \rangle})^T\right ) \\
				=& \left (\Re(\boldsymbol{\{\mathcal{Z}}^{(k)} \}_{ {k=1}}^n )_{\left \langle n \right \rangle} \odot_1 \Re(\{\boldsymbol{\mathcal{G}}^{(k)} \}_{ {k=1}}^n )_{\left \langle n \right \rangle}\right ) \\
				&\left (\Re( \{\boldsymbol{\mathcal{Z}}^{(k)} \}_{ {k=n+1}}^N )_{\left \langle N-n \right \rangle} \odot_1 \Re( \{\boldsymbol{\mathcal{G}}^{(k)} \}_{ {k=n+1}}^N )_{\left \langle N-n \right \rangle}\right ) ^T,
				\label{proof1.1}
		\end{split}\end{equation}
		we can futher obtain
		\begin{equation}\begin{split}
				(\boldsymbol{\mathcal{A}}\circledast \boldsymbol{\mathcal{B}})_{(n)}=& (\boldsymbol{\mathcal{Z}}^{(n)}\odot_n \boldsymbol{\mathcal{G}}^{(n)})_{(n)} (\boldsymbol{\mathcal{Z}}^{(\neq n)}_{\left \langle N-1 \right \rangle} \odot_1 \boldsymbol{\mathcal{G}}^{(\neq n)}_{\left \langle N-1 \right \rangle})^T\\
				=&\boldsymbol{\mathcal{Y}}^{(n)}_{(n)}(\boldsymbol{\mathcal{Z}}^{(\neq n)}_{\left \langle N-1 \right \rangle} \odot_1 \boldsymbol{\mathcal{G}}^{(\neq n)}_{\left \langle N-1 \right \rangle})^T,\\
				\boldsymbol{{Z}}^{(\leq n)}_{\left \langle n \right \rangle}\odot_1\boldsymbol{{g}}^{(\leq n)}_{\left \langle n \right \rangle}=&\Re(\Re(\{\boldsymbol{\mathcal{Z}}^{(k)} \}_{ {k=1}}^{n-1}), \boldsymbol{\mathcal{Z}}^{( n)})_{\left \langle n \right \rangle} \odot_1\\
				&\left (\Re(\Re(\{\boldsymbol{\mathcal{G}}^{(k)} \}_{ {k=1}}^{n-1}), \boldsymbol{\mathcal{G}}^{( n)})_{\left \langle n \right \rangle}\right )^T\\
				=& \Re(\Re(\{\boldsymbol{\mathcal{Z}}^{(k)} \}_{ {k=1}}^{n-1})_{\left \langle n-1 \right \rangle} \odot_1\\ &\Re(\{\boldsymbol{\mathcal{G}}^{(k)} \}_{ {k=1}}^{n-1})_{\left \langle n-1 \right \rangle}), \boldsymbol{\mathcal{Y}}^{(n)}\\
		\end{split}\end{equation}
		Therefore, Eq.(\ref{eq18}) holds by mathematical induction and the Hadamard product of tensors with TN representation can perform in their factors via mode-n  Khatri-Rao product.
		
		\it
		\textbf{2)} The inner product of $\boldsymbol{\mathcal{A}}$ and $\boldsymbol{\mathcal{B}}$ can be computed by
		\begin{equation}\begin{split}
				\left \langle \boldsymbol{\mathcal{A}}, \boldsymbol{\mathcal{B}}  \right \rangle &= \left \langle \boldsymbol{\mathcal{A}}\circledast \boldsymbol{\mathcal{B}}, \circ_{k=1}^N \mu_k \right \rangle\\
				&=(\boldsymbol{\mathcal{A}}\circledast \boldsymbol{\mathcal{B}})\times_1 u_1^T\times_2\cdots\times_N u_N^T\\
				&=\Re(\boldsymbol{\mathcal{Y}}^{(1)},\cdots, \boldsymbol{\mathcal{Y}}^{(N)})\times_1 u_1^T\times_2\cdots\times_N u_N^T\\
				&=\Re(\boldsymbol{\mathcal{Y}}^{(1)}\times_1 u_1^T, \cdots, \boldsymbol{\mathcal{Y}}^{(N)}\times_N u_N^T),
				\label{eq19}
		\end{split}\end{equation}
		where $u_k=(1,\cdots,1)\in \mathbb{R} ^{I_k}$ is a vector with all values of one. Naturally, the Frobenius norm of $\boldsymbol{\mathcal{A}}$ can be obtained by $\sqrt{\left \langle \boldsymbol{\mathcal{A}}, \boldsymbol{\mathcal{A}}  \right \rangle} $.
	\end{Proposition}
	
	\begin{Proposition}\it
		Let an $N$-order tensor $\boldsymbol{\mathcal{A}}\in \mathbb{R} ^{I_1\times \cdots \times I_N}$ and $M$th-order tensor $\boldsymbol{\mathcal{B}}\in \mathbb{R} ^{J_1\times \cdots \times J_M}$ be respectively decomposed by $\Psi_1$ and $\Psi_2$, which have the TN representations of $\boldsymbol{\mathcal{A}}=\Re(\boldsymbol{\mathcal{Z}}^{(1)},\cdots, \boldsymbol{\mathcal{Z}}^{(N)} )$ and $\boldsymbol{\mathcal{B}}=\Re(\boldsymbol{\mathcal{G}}^{(1)},\cdots, \boldsymbol{\mathcal{G}}^{(M)} )$. The outer product (see Def.~\ref{def2.6}) of $\boldsymbol{\mathcal{A}}$ and $\boldsymbol{\mathcal{B}}$ yields a tensor $\boldsymbol{\mathcal{C}}=\boldsymbol{\mathcal{A}}\circ \boldsymbol{\mathcal{B}}$, which has a TN representation of $\boldsymbol{\mathcal{C}}=\Re(\boldsymbol{\mathcal{Z}}^{(1)},\cdots, \boldsymbol{\mathcal{Z}}^{(N)},\boldsymbol{\mathcal{G}}^{(1)},\cdots, \boldsymbol{\mathcal{G}}^{(M)})$.
	\end{Proposition}
	
	Moreover, the Kronecker and mode-n Khatri-Rao product of tensors with TN representation can perform in their factors and the result will also follow the TN format.

	\begin{Proposition}
		\label{th5}
		\it
		Let two $N$-order tensors $\boldsymbol{\mathcal{A}}\in \mathbb{R} ^{I_1\times \cdots \times I_N}$ and $\boldsymbol{\mathcal{B}}\in \mathbb{R} ^{J_1\times \cdots \times J_N}$ be decomposed by $\Psi_1$ and $\Psi_2$, which has a TN representations of $\boldsymbol{\mathcal{A}}=\Re(\boldsymbol{\mathcal{Z}}^{(1)},\cdots, \boldsymbol{\mathcal{Z}}^{(N)} )$ and $\boldsymbol{\mathcal{B}}=\Re(\boldsymbol{\mathcal{G}}^{(1)},\cdots, \boldsymbol{\mathcal{G}}^{(N)} )$, where $\boldsymbol{\mathcal{Z}}^{(k)}\in \mathbb{R} ^{R_{1,k}\times \dots \times I_k \times \dots\times R_{N,k}}$ and $\boldsymbol{\mathcal{G}}^{(k)}\in \mathbb{R} ^{S_{1,k}\times \dots \times J_k \times \dots\times S_{N,k}}$ for $k=1,\cdots, N$. The Kronecker product (see Def.~\ref{def2.4}) of $\boldsymbol{\mathcal{A}}$ and $\boldsymbol{\mathcal{B}}$ yields a tensor $\boldsymbol{\mathcal{C}}=\boldsymbol{\mathcal{A}}\otimes \boldsymbol{\mathcal{B}}$, having the TN representation of $\boldsymbol{\mathcal{C}}=\Re(\boldsymbol{\mathcal{Y}}^{(1)},\cdots, \boldsymbol{\mathcal{Y}}^{(N)})$, where each factor $\boldsymbol{\mathcal{Y}}^{(k)}$ of size $R_{1,k}S_{1,k}\times \dots \times I_k^2 \times \dots\times R_{N,k}S_{N,k} $, for $k=1,\cdots,N$, can be computed by
		\begin{equation}\begin{split}
				\boldsymbol{\mathcal{Y}}^{(k)}=\boldsymbol{\mathcal{Z}}^{(k)}\otimes \boldsymbol{\mathcal{G}}^{(k)}.
				\label{eq20}
		\end{split}\end{equation}
	\end{Proposition}
	Proof. The Kronecker product of $\boldsymbol{\mathcal{A}}$ and $\boldsymbol{\mathcal{B}}$ can be written as
	\begin{equation}\begin{split}
			\boldsymbol{\mathcal{A}} \otimes \boldsymbol{\mathcal{B}}=&\Re(\boldsymbol{\mathcal{Z}}^{(1)},\cdots, \boldsymbol{\mathcal{Z}}^{(N)} )\otimes \Re(\boldsymbol{\mathcal{G}}^{(1)},\cdots, \boldsymbol{\mathcal{G}}^{(N)} )\\
			=& \Re(\boldsymbol{\mathcal{Z}}^{(1)}\otimes \boldsymbol{\mathcal{G}}^{(1)}, \boldsymbol{\mathcal{Z}}^{(\geq 2)}\otimes \boldsymbol{\mathcal{G}}^{(\geq 2)})\\
			=& \Re(\boldsymbol{\mathcal{Z}}^{(1)}\otimes \boldsymbol{\mathcal{G}}^{(1)}, \boldsymbol{\mathcal{Z}}^{(2)}\otimes \boldsymbol{\mathcal{G}}^{(2)}, \boldsymbol{\mathcal{Z}}^{(\geq 3)}\otimes \boldsymbol{\mathcal{G}}^{(\geq 3)})\\
			=&\cdots= \Re(\{\boldsymbol{\mathcal{Z}}^{(k)}\otimes \boldsymbol{\mathcal{G}}^{(k)} \}_{k=1}^N)
			\label{proof2.1}
	\end{split}\end{equation}

	\begin{Proposition}
		\label{thrao}
		\it
		Let two $N$-order tensors $\boldsymbol{\mathcal{A}}\in \mathbb{R} ^{I_1\times \cdots \times I_N}$ and $\boldsymbol{\mathcal{B}}\in \mathbb{R} ^{J_1\times \cdots \times J_N}$ be decomposed by $\Psi_1$ and $\Psi_2$, which have the TN representations of $\boldsymbol{\mathcal{A}}=\Re(\boldsymbol{\mathcal{Z}}^{(1)},\cdots, \boldsymbol{\mathcal{Z}}^{(N)} )$ and $\boldsymbol{\mathcal{B}}=\Re(\boldsymbol{\mathcal{G}}^{(1)},\cdots, \boldsymbol{\mathcal{G}}^{(N)} )$, where $\boldsymbol{\mathcal{Z}}^{(k)}\in \mathbb{R} ^{R_{1,k}\times \dots \times I_k \times \dots\times R_{N,k}}$ and $\boldsymbol{\mathcal{G}}^{(k)}\in \mathbb{R} ^{S_{1,k}\times \dots \times J_k \times \dots\times S_{N,k}}$ for $k=1,\cdots, N$. Suppose that $I_n=J_n, n\in [N]$, then the mode-n Khatri-Rao product (see Def.~\ref{def2.5}) between these two tensors yields a tensor $\boldsymbol{\mathcal{C}}=\boldsymbol{\mathcal{A}}\odot_{n} \boldsymbol{\mathcal{B}}$, having a TN representation of $\boldsymbol{\mathcal{C}}=\Re(\boldsymbol{\mathcal{Y}}^{(1)},\cdots, \boldsymbol{\mathcal{Y}}^{(N)})$, where each factor $\boldsymbol{\mathcal{Y}}^{(k)}$, for $k=1,\cdots,N$, can be computed by
		\begin{equation}\begin{split}
				\boldsymbol{\mathcal{Y}}^{(k)}=
				\left\{\begin{matrix}
					\boldsymbol{\mathcal{Z}}^{(k)}\odot_{k} \boldsymbol{\mathcal{G}}^{(k)} &  k=n,\\
					\boldsymbol{\mathcal{Z}}^{(k)}\otimes_k \boldsymbol{\mathcal{G}}^{(k)} & k\neq n.
				\end{matrix}\right.,
				\label{eqrao}
		\end{split}\end{equation}
	\end{Proposition}

	The Propositions \ref{th2}$\sim$\ref{thrao} establishes the essential multilinear operation of tensors with TN representation, which is implemented straightly on the latent space and obtain the TN format results. Note that these operations may cause changes in the size of the factors and aggravate the computational complexity of the tensor contraction. We further define the addition of the tensors with TN representation in Theorem~\ref{th6}, indicating that the combination of multiple TD models essentially can be transformed into a single one.

	\begin{Theorem}
		\label{th6}
		\it
		Let two $N$-order tensors $\boldsymbol{\mathcal{A}}, \boldsymbol{\mathcal{B}}\in \mathbb{R} ^{I_1\times \cdots \times I_N}$ be decomposed by $\Psi_1$ and $\Psi_2$, which have the TN representations of $\boldsymbol{\mathcal{A}}=\Re(\boldsymbol{\mathcal{Z}}^{(1)},\cdots, \boldsymbol{\mathcal{Z}}^{(N)} )$ and $\boldsymbol{\mathcal{B}}=\Re(\boldsymbol{\mathcal{G}}^{(1)},\cdots, \boldsymbol{\mathcal{G}}^{(N)} )$, where $\boldsymbol{\mathcal{Z}}^{(k)}\in \mathbb{R} ^{R_{1,k}\times \dots \times I_k \times \dots\times R_{N,k}}$ and $\boldsymbol{\mathcal{G}}^{(k)}\in \mathbb{R} ^{S_{1,k}\times \dots \times I_k \times \dots\times S_{N,k}}$ for $k=1,\cdots, N$. Then the addition of these two tensors, $\boldsymbol{\mathcal{C}}=\boldsymbol{\mathcal{A}}+\boldsymbol{\mathcal{B}}$, having a TN representation of $\boldsymbol{\mathcal{C}}=\Re(\boldsymbol{\mathcal{Y}}^{(1)},\cdots,\boldsymbol{\mathcal{Y}}^{(N)})$, where each factor $\boldsymbol{\mathcal{Y}}^{(k)}$ of size ${P_{1,k}\times \dots \times I_k \times \dots\times P_{N,k}}$, for $k=1,\cdots,N$, can be computed by
		\begin{small}
			\begin{equation}\begin{split}
					&P_{i,j}=
					\left\{\begin{matrix}
						1&  R_{i,j}=S_{i,j}=1,\\
						R_{i,j}+S_{i,j}&others, 
					\end{matrix}\right. 1\leq i<j\leq N,\\
					&\boldsymbol{\mathcal{Y}}^{(k)}_{p_{1,k},\cdots,i_k,\cdots,p_{N,k}}=\left\{\begin{matrix}
						\boldsymbol{\mathcal{Z}}^{(k)}_{p_{1,k},\cdots,i_k,\cdots,p_{N,k}}\\
						\qquad\qquad (1\leq p_{i,k}\leq R_{i,k}),\\
						\boldsymbol{\mathcal{G}}^{(k)}_{p_{1,k}-R_{i,k},\cdots,i_k,\cdots,p_{N,k}-R_{N,k}}\\
						\qquad\qquad (R_{i,k}<p_{i,k}\leq P_{i,k}),\\
						0 \qquad\quad others.
					\end{matrix}\right.
					\label{eq21}
			\end{split}\end{equation}
		\end{small}
	\end{Theorem}

	The establishment of Theorem~\ref{th6} requires that the topology of the resulting tensor TN representation cannot contain isolated factors; otherwise we will randomly construct an additional edge for the factors with zero connections. It can be observed that the addition of tensors with TN representations is the combination of the corresponding factors into a higher-order diagonalized format. Therefore, we have the following remark:

	\begin{Remark}\it
$ $

\begin{enumerate}[(1)]
	\item The addition of tensors represented by the same TN models without isolated vertices can obtain a tensor in a TN format with identical topology.
	\item At least addition $\frac{N-1}{2}$ (resp. $\frac{N}{2}$) tensors with TR (resp. TT) representation to guarantee that the result tensor corresponds topology is a complete graph.
\end{enumerate}
	\end{Remark}

	In summary, multilinear operations among higher-order tensors can be carried out implicitly in the latent space, and the storage cost is linear with the tensor order.

	\section{MTNR FOR LOW-RANK TENSOR COMPLETION}
	\label{sec4}
	
	This section applies MTNR to the LRTC task, aiming to obtain a low-rank approximation of the data from limited samplings. We propose two practical algorithms, named MTNR-ALS and MTNR-ADMM, and their computational complexity is analyzed.
	
	\subsection{MTNR-ALS Algorithm}
	The MTNR-ALS is an extension of Algorithm 1 on the LRTC task. Given an $N$-order incomplete observation tensor $\boldsymbol{\mathcal{M}}\in \mathbb{R} ^{I_1\times \cdots \times I_N}$ and entries index set $\Omega$. Our objective is to obtain the approximated tensor, $\boldsymbol{\mathcal{X}}$, in the MTNR format (\ref{eq10}). Hence, the MTNR-based optimizing model for LRTC can be formulated as
	\begin{equation}\begin{split}
			\min_{\boldsymbol{\mathcal{X}}, \boldsymbol{\mathcal{Z}}^{(1)}_1,\cdots,\boldsymbol{\mathcal{Z}}^{(N)}_r}&\quad ||\boldsymbol{\mathcal{X}}- \sum_{i=k}^r \Re(\boldsymbol{\mathcal{Z}}^{(1)}_k,\cdots, \boldsymbol{\mathcal{Z}}^{(N)}_k)||_F^2\\
			\text{s.t.}\ \  &P_{\Omega}(\boldsymbol{\mathcal{X}})= P_{\Omega}(\boldsymbol{\mathcal{M}}),
			\label{eq22}
	\end{split}\end{equation}
	where $\Re(\boldsymbol{\mathcal{Z}}_k^{(1)},\cdots,\boldsymbol{\mathcal{Z}}_k^{(N)})$ represents the $k$th low-rank component $\boldsymbol{\mathcal{A}}_{k}$. For the sake of clarity, we omit the constraint condition in (\ref{eq11}).  The (\ref{eq22}) cannot be optimized directly due to $r $ is assumed to be an unknown coefficient. Consistent with (\ref{eq12}), we consider the $k$th low-rank component's optimization model
	\begin{equation}\begin{split}
			\min_{\boldsymbol{\mathcal{T}}, \boldsymbol{\mathcal{Z}}^{(1)}_k,\cdots,\boldsymbol{\mathcal{Z}}^{(N)}_k}&\ \ ||\mathcal{T} - \Re(\boldsymbol{\mathcal{Z}}^{(1)}_k,\cdots, \boldsymbol{\mathcal{Z}}^{(N)}_k)||_F^2\\
			\text{s.t.}\ \  &P_{\Omega}(\boldsymbol{\mathcal{T}}) = P_{\Omega}(\boldsymbol{\mathcal{M}} - \sum_{i=1}^{k-1} \boldsymbol{\mathcal{A}}_i).
			\label{eq23}
	\end{split}\end{equation}
	We use the $\boldsymbol{\mathcal{A}}_{k} $ to fit the present residual observation tensor $\boldsymbol{\mathcal{M}} - \sum_{i=1}^{k-1} \boldsymbol{\mathcal{A}}_i$. The variables $\boldsymbol {\mathcal{T}}$ and $\boldsymbol{\mathcal{Z}}_k^{(i)}$ are interrelated. We employ the ALS scheme to address (\ref{eq23}) by alternately updating
	\begin{small}
		\begin{equation}\begin{split}
				\left\{\begin{array}{l}
					\boldsymbol{\mathcal{Z}}^{(n)}_{k} = \mathop{\arg\min}\limits_{\boldsymbol{\mathcal{Z}}^{(n)}_k}\  || \boldsymbol{{T}}_{(n)} - \boldsymbol{{Z}}^{(n)}_{k(n)} (\boldsymbol{{Z}}^{(\neq n)}_{k\left \langle N-1 \right \rangle })^T ||_F^2, n=1,\cdots,N,
					\\
					\boldsymbol{\mathcal{T}}= P_{\Omega^c}(\Re(\boldsymbol{\mathcal{Z}}^{(1)}_k,\cdots, \boldsymbol{\mathcal{Z}}^{(N)}_k)) + P_{\Omega}(\boldsymbol{\mathcal{M}} - \sum_{i=1}^{k-1}\boldsymbol{\mathcal{A}}_i),
				\end{array}\right.
				\label{eq24}
		\end{split}\end{equation}
	\end{small}
	
	\noindent where $\Omega^c$ denotes indices set of missing entries, and the closed-form solution of $\boldsymbol{\mathcal{Z}}^{(n)}_{k}$-subproblem is given in (\ref{eq14}). We employ the same strategy as the ATL algorithm to perform the update of the factor structure. The optimizing model (\ref{eq22}) can be regarded as MTNR if the observed data is no missing elements. The whole implementation process of the MTNR-ALS algorithm is listed in Algorithm \ref{alg2}.
	
	Compared with other TD-based approaches for LRTC, e.g., TT-ALS and TR-ALS~\cite{ref60}, the main advantage of MTNR-ALS is the TN representation with adaptive topology and the number of low-rank components is adjusted dynamically, which is helpful for capturing the low-rankness inside the high-dimension data. Nevertheless, the low-rankness of the recovered tensor is weakened during the increase of the number of low-rank components, which may be unfavorable for the results. We consider developing a new algorithm to guarantee the low-rank property of the reconstructed tensor.
	
	\floatname{algorithm}{Algorithm}
	\renewcommand{\algorithmicrequire}{\textbf{Input:}}  
	\renewcommand{\algorithmicensure}{\textbf{Output:}}    
	\begin{algorithm} [t]
		\caption{MTNR-ALS Algorithm for Tensor Completion} 
		\begin{algorithmic}[1]
			\label{alg2}
			\REQUIRE an $N$-order incomplete tensor $\boldsymbol{\mathcal{M}}\in \mathbb{R} ^{I_1\times \dots \times I_N}$, observed entries index set $\Omega$, relative error $\epsilon$, complexity restrain $\gamma, S_{max}$, and threshold $\delta$.
			\ENSURE Recovered tensor $\boldsymbol{{\mathcal{X}}}$.
			\STATE $\boldsymbol{\mathcal{X}}=P_{\Omega}(\boldsymbol{\mathcal{M}}),k\gets1$.
			\REPEAT
			\STATE Initialize $R_{(i,j)}^k\gets 1 $ for $ 1\leq i < j \leq N$, $\boldsymbol{\mathcal{Z}}_k^{(n)} \in \mathbb{R} ^{R_{(1,n)}^k\times\cdots R_{(n-1,n)}^k \times I_n\times R_{(n,n+1)}^k \dots \times R_{(n,N)}^k}$ for $n=1,\cdots,N$.
			
			\FOR{$s$=1 to $S_{max}$}
			\STATE $\boldsymbol{\mathcal{A}}_k\gets \Re(\boldsymbol{\mathcal{Z}}_k^{(1)},\cdots, \boldsymbol{\mathcal{Z}}_k^{(N)})$.
			\STATE Update $\boldsymbol{\mathcal{Z}}_k^{(n)}$ via (\ref{eq14}), $n=1,\cdots, N$.
			\STATE Update $\boldsymbol{{\mathcal{T}}}$ via (\ref{eq24}).
			\STATE Steps (7-13) in algorithm~\ref{alg1}.
			\ENDFOR
			\STATE $\boldsymbol{{\mathcal{X}}}\gets \boldsymbol{{\mathcal{X}}} + P_{\Omega^c}(\boldsymbol{\mathcal{A}}_k), k\gets k+1$.
			\UNTIL The convergence condition $\frac{|| P_{\Omega}(\boldsymbol{\mathcal{A}}_k) ||_F}{||P_{\Omega} (\boldsymbol{\mathcal{M}})||_F} < \epsilon$ is achieved.
		\end{algorithmic}
	\end{algorithm}
	
	\subsection{MTNR-ADMM Algorithm}
	\renewcommand{\baselinestretch}{1.2}
	The TD-based LRTC methods incorporating low-rank regularizations can achieve better performance~\cite{ref43} as the explicit tensor rank constraint and the implicit convex surrogate function (e.g., SNN~\cite{ref5}) are applied to the recovered tensor to better capture the global low-rank structure information. However, the computational complexity of SNN that directly acts on the recovery tensor is $\mathcal{O}(NI^{N+1})$, which grows exponentially with the data order and obtains suboptimal results. Thereby, we consider applying implicit low-rank regularizations to the factors of TN representation to reduce the calculation. In a nutshell, we impose SNN regularization to the factors instead of the original format with the computational complexity $\mathcal{O}(NIR^{t}(Rt+I))$, which is linear to tensor order $N$. Specifically, the optimization model of $k$th low-rank component is reformulated as

	\begin{equation}\begin{split}
		\resizebox{0.85\hsize}{!}{$\begin{aligned}
				\min_{\boldsymbol{\mathcal{T}}, [\boldsymbol{\mathcal{Z}}]}&\ \ \sum_{i=1}^N \sum_{n=1}^N ||\boldsymbol{{Z}}^{(i)}_{k(n)}||_* + \frac{\lambda}{2}||\mathcal{T} - \Re(\boldsymbol{\mathcal{Z}}^{(1)}_k,\cdots, \boldsymbol{\mathcal{Z}}^{(N)}_k)||_F^2,\\
				& \text{s.t.}\ \  P_{\Omega}(\boldsymbol{\mathcal{T}}) = P_{\Omega}(\boldsymbol{\mathcal{M}}- \sum_{i=1}^{k-1} \boldsymbol{\mathcal{A}}_i),
				\label{eq26}
		\end{aligned}$}
	\end{split}\end{equation}
	
	\noindent where $\lambda>0$ is a trade-off coefficient. Note that the low-rank regularization is dynamically adjusted as the factor structure update during the optimization process. If a mode-$n$ matricization of $\boldsymbol{\mathcal{Z}}^{(i)}_{k}$ is present in vector format, then the corresponding nuclear norm will be invalid. Since variables in (\ref{eq26}) are inter-coupled and cannot optimized independently, we introduce auxiliary variables $\{\boldsymbol{\mathcal{G}}_{n}^{(i)}\}_{i=1,n=1}^{N ,N}$ to reduce the optimization difficulty and the model (\ref{eq26}) is deduced to
	\begin{equation}\begin{split}
            \resizebox{0.85\hsize}{!}{$\begin{aligned}
			\min_{\boldsymbol{\mathcal{T}}, [\boldsymbol{\mathcal{Z}}],[\boldsymbol{\mathcal{G}}]}&\ \ \sum_{i=1}^N \sum_{n=1}^N ||\boldsymbol{{g}}_{n(n)}^{(i)}||_*  + \frac{\lambda}{2}||\mathcal{T} - \Re(\boldsymbol{\mathcal{Z}}^{(1)}_k,\cdots, \boldsymbol{\mathcal{Z}}^{(N)}_k)||_F^2,\\
			& \text{s.t.}\ \ \boldsymbol{\mathcal{G}}_{n}^{(i)}=\boldsymbol{\mathcal{Z}}^{(i)}_{k},i=1,\cdots,N,n=1,\cdots,N,\\
			&\quad\quad\quad P_{\Omega}(\boldsymbol{\mathcal{T}}) = P_{\Omega}(\boldsymbol{\mathcal{M}}- \sum_{i=1}^{k-1} \boldsymbol{\mathcal{A}}_i),
			\label{eq27}
			\end{aligned}$}
	\end{split}\end{equation}
	where $[\boldsymbol{\mathcal{Z}}]$ and $[\boldsymbol{\mathcal{G}}]$ represent $\{\boldsymbol{\mathcal{Z}}^{(i)}_k\}_{i=1}^N$ and $\{\boldsymbol{\mathcal{G}}_{n}^{(i)}\}_{i=1,n=1}^{N ,N}$, respectively. The above multivariable optimization problem can be solved via the ADMM framework. The augmented Lagrangian function of (\ref{eq27}) can be formulated as follows
	\begin{equation}\begin{split}
			\boldsymbol{L}&(\boldsymbol{\mathcal{T}},[\boldsymbol{\mathcal{Z}}],[\boldsymbol{\mathcal{G}}], [\boldsymbol{\mathcal{Y}}])=\sum_{i=1}^N \sum_{n=1}^N (||\boldsymbol{{g}}_{n(n)}^{(i)}||_*  + \frac{\rho}{2} ||\boldsymbol{\mathcal{G}}_{n}^{(i)} - \boldsymbol{\mathcal{Z}}^{(i)}_{k}||_F^2 + \\
			&< \boldsymbol{\mathcal{Y}}_{n}^{(i)}, \boldsymbol{\mathcal{G}}_{n}^{(i)} - \boldsymbol{\mathcal{Z}}^{(i)}_{k} > ) + \frac{\lambda}{2}||\mathcal{T} - \Re(\boldsymbol{\mathcal{Z}}^{(1)}_k,\cdots, \boldsymbol{\mathcal{Z}}^{(N)}_k)||_F^2,\\
			&\ \ \ \ \ \ \ \ \ \ \ \ \  \text{s.t.}\quad  P_{\Omega}(\boldsymbol{\mathcal{T}}) = P_{\Omega}(\boldsymbol{\mathcal{M}}- \sum_{i=1}^{k-1} \boldsymbol{\mathcal{A}}_i).
			\label{eq28}
	\end{split}\end{equation}
	where $[\boldsymbol{\mathcal{Y}}]=\{\boldsymbol{\mathcal{Y}}_{n}^{(i)}\}_{i=1,n=1}^{N,N}$ is Lagrangian multiplier tensor set and $\rho$ is a positive penalty parameter. The solution of (\ref{eq28}) is given via alternately updating variables $\boldsymbol{\mathcal{T}},[\boldsymbol{\mathcal{Z}}],[\boldsymbol{\mathcal{G}}]$ and $ [\boldsymbol{\mathcal {Y}}]$. The update mechanism is to optimize one variable each time and fix others, which is as follows:
	\begin{enumerate}[(1)]
		\item \textbf{Update} $[\boldsymbol{\mathcal{Z}}]$. Extract all items that are associated with $\boldsymbol{\mathcal{Z}}$ from (\ref{eq28}). Then the $\boldsymbol{\mathcal{Z}}^{(i)}_{k}$-subproblem, for $i=1,\cdots,N$, can be formulated as:
		\begin{equation}\begin{split}
				\min_{\boldsymbol{\mathcal{Z}}^{(i)}_{k}}\ \ \sum_{n=1}^N \frac{\rho}{2} ||&\boldsymbol{\mathcal{G}}_{n}^{(i)} -\boldsymbol{\mathcal{Z}}^{(i)}_{k} + \frac{1}{\rho} \boldsymbol{\mathcal{Y}}_{n}^{(i)}||_F^2 + \\
				&\frac{\lambda}{2}||\mathcal{T} - \Re(\boldsymbol{\mathcal{Z}}^{(1)}_k,\cdots, \boldsymbol{\mathcal{Z}}^{(N)}_k)||_F^2.
				\label{eq29}
		\end{split}\end{equation}
		The closed-form solution of $\boldsymbol{\mathcal{Z}}^{(i)}_{k}$ is given in mode-$i$ matricization format as follows:
		\begin{equation}\begin{split}
				\boldsymbol{{Z}}^{(i)}_{k(i)}= ( (\sum_{n=1}^N \rho & \boldsymbol{\mathcal{G}}_{n}^{(i)} + \boldsymbol{\mathcal{Y}}_{n}^{(i)})_{(i)}+ \lambda \boldsymbol{{T}}_{(i)} \boldsymbol{{Z}}^{\neq i}_{k{\left \langle N-1 \right \rangle}})  \\
				&(\lambda  (\boldsymbol{{Z}}^{(\neq i)}_{k \left \langle N-1 \right \rangle})^T \boldsymbol{{Z}}^{(\neq i)}_{k \left \langle N-1 \right \rangle} + \rho N I)^{-1},
				\label{eq30}
		\end{split}\end{equation}
		where $I$ is a identity matrix of size $\prod_{^{n=1}_{n\neq i}}^{N} R_{n,i}\times \prod_{^{n=1}_{n\neq i}}^{N} R_{n,i} $.
		
		\item \textbf{Update} $[\boldsymbol{\mathcal{G}}]$. The optimization subprobelem for $\boldsymbol{\mathcal{G}}_n^{(i)}$,  $i=1\cdots, N,n=1\cdots, N$, is formulated to
		\begin{equation}\begin{split}
				&\min_{\boldsymbol{{g}}_n^{(i)}}\ \  ||\boldsymbol{\mathcal{G}}_{n(i)}^{(i)}||_* + \frac{\rho}{2} ||\boldsymbol{\mathcal{G}}_{n}^{(i)} - \boldsymbol{\mathcal{Z}}^{(i)}_{k}+ \frac{1}{\rho} \boldsymbol{\mathcal{Y}}_{n}^{(i)}||_F^2
				\label{eq31}
		\end{split}\end{equation}
		The closed-form solution of above model is given in the mode-$i$ matricization format as follows:
		\begin{equation}\begin{split}
				\boldsymbol{\mathcal{G}}_{n(i)}^{(i)} = \mathcal{D}_{\frac{1}{\rho}}((\boldsymbol{\mathcal{Z}}^{(i)}_{k}- \frac{1}{\rho} \boldsymbol{\mathcal{Y}}_{n}^{(i)})_{(i)}).
				\label{eq32}
		\end{split}\end{equation}
		where $\mathcal{D}_{\mu}(\cdot)$ is the singular value thresholding (SVT)~\cite{ek2012singular} operator. Given a matrix $A$, whoseich singular value decomposition is expressed as $USV^t$, then $\mathcal{D}_ {\mu}(A)=U\max\{S-\mu I, 0 \}V^T$, where $I$ is the identity matrix of the same size as $A$.
		
		\item \textbf{Update} $[\boldsymbol{\mathcal{Y}}]$ and $\boldsymbol{\mathcal{T}}$. The update rules w.r.t $\boldsymbol{\mathcal{Y}}_n^i$, for $i,n\in [N]$, is 
		\begin{equation}\begin{split}
				\boldsymbol{\mathcal{Y}}_{n}^{(i)} = \boldsymbol{\mathcal{Y}}_{n}^{(i)} + \rho(\boldsymbol{\mathcal{G}}_{n}^{(i)} - \boldsymbol{\mathcal{Z}}^{(i)}_{k}) .
				\label{eq33}
		\end{split}\end{equation}
		The update strategy of $\boldsymbol{\mathcal{T}}$ is consistent with (\ref{eq24}), which can be computed by
		\begin{equation}\begin{split}
				\boldsymbol{\mathcal{T}}= P_{\Omega^c}(\Re(\boldsymbol{\mathcal{Z}}^{(1)}_k,\cdots, \boldsymbol{\mathcal{Z}}^{(N)}_k) + P_{\Omega}(\boldsymbol{\mathcal{M}} - \sum_{i=1}^{k-1}\boldsymbol{\mathcal{A}}_i).
				\label{eq34}
		\end{split}\end{equation}
	\end{enumerate}
	Algorithm~\ref{alg3} summarizes the overall optimization steps of MTNR-ADMM.
	
	\floatname{algorithm}{Algorithm}
	\renewcommand{\algorithmicrequire}{\textbf{Input:}}  
	\renewcommand{\algorithmicensure}{\textbf{Output:}}    
	\begin{algorithm} [t]
		\caption{MTNR-ADMM Algorithm for Tensor Completion} 
		\begin{algorithmic}[1]
			\label{alg3}
			\REQUIRE an $N$-order incomplete tensor $\boldsymbol{\mathcal{M}}\in \mathbb{R} ^{I_1\times \dots \times I_N}$, observed entries set $\Omega$, relative error $\epsilon$, complexity restrain $\gamma, S_{max}$, threshold $\delta$, trade-off coefficient $\lambda$ and penalty parameter $\rho, \rho_{\max}$.
			\ENSURE Recovered tensor $\boldsymbol{{\mathcal{X}}}$.
			\STATE $\boldsymbol{\mathcal{X}}=P_{\Omega}(\boldsymbol{\mathcal{M}}),k\gets1$.
			\REPEAT
			\STATE Initialize $R_{(i,j)}^k\gets 1 $ for $ 1\leq i < j \leq N$, $\boldsymbol{\mathcal{Z}}_k^{(n)} \in \mathbb{R} ^{R_{(1,n)}^k\times\cdots R_{(n-1,n)}^k \times I_n\times R_{(n,n+1)}^k \dots \times R_{(n,N)}^k}$ for $n=1,\cdots,N$, $\{\boldsymbol{\mathcal{G}}_{n}^{(i)}\}_{i=1,n=1}^{N,N}$ and $\{\boldsymbol{\mathcal{Y}}_{n}^{(i)}\}_{i=1,n=1}^{N,N} $.
			
			\FOR{$s$=1 to $S_{max}$}
			\STATE $\boldsymbol{\mathcal{A}}_k\gets \Re(\boldsymbol{\mathcal{Z}}_k^{(1)},\cdots, \boldsymbol{\mathcal{Z}}_k^{(N)})$.
			\STATE Update $\boldsymbol{\mathcal{Z}}_k^{(n)}$ via (\ref{eq30}), $n=1,\cdots, N$.
			\STATE Update $\boldsymbol{\mathcal{G}}_{n}^{(i)}$ via (\ref{eq32}), $i=1,\cdots, N, n=1,\cdots, N$.
			\STATE Update $\boldsymbol{\mathcal{Y}}_{n}^{(i)}$ via (\ref{eq33}), $i=1,\cdots, N, n=1,\cdots, N$.
			\STATE Update $\boldsymbol{{\mathcal{T}}}$ via (\ref{eq24}).
			\STATE $\rho\gets \min(1.01\rho, \rho_{\max})$.
			
			\STATE $rse\gets \frac{||\Re(\boldsymbol{\mathcal{Z}}_k^{(1)},\cdots, \boldsymbol{\mathcal{Z}}_k^{(N)})  - \boldsymbol{\mathcal{A}}_k ||_F}{||\boldsymbol{\mathcal{A}}_k||_F}$
			
			\IF{$rse \leq \delta$ and ${\mathcal{P} }(\boldsymbol{\mathcal{Z}}_k^{(1)},\cdots, \boldsymbol{\mathcal{Z}}_k^{(N)})\leq\gamma$}
			\STATE Select factors pair $(\boldsymbol{\mathcal{Z}}_k^{(i)}, \boldsymbol{\mathcal{Z}}_k^{(j)}) $ via (\ref{eq15}).
			\STATE $R_{(i,j)}^k\gets R_{(i,j)}^k+1$.
			\STATE Add a new slice to $j$th mode of $\boldsymbol{\mathcal{Z}}_k^{(i)}$, $\{\boldsymbol{\mathcal{G}}_{n}^{(i)}\}_{n=1}^{N} $ and $\{\boldsymbol{\mathcal{Y}}_{n}^{(i)}\}_{n=1}^{N}$.
			\STATE Add a new slice to $i$th mode of $\boldsymbol{\mathcal{Z}}_k^{(j)}$, $\{\boldsymbol{\mathcal{G}}_{n}^{(j)}\}_{n=1}^{N} $ and $\{\boldsymbol{\mathcal{Y}}_{n}^{(j)}\}_{n=1}^{N}$.
			\ENDIF 
			
			\ENDFOR
			\STATE $\boldsymbol{{\mathcal{X}}}\gets \boldsymbol{{\mathcal{X}}} + P_{\Omega^c}(\boldsymbol{\mathcal{A}}_k), k\gets k+1$.
			\UNTIL The convergence condition $\frac{|| P_{\Omega}(\boldsymbol{\mathcal{A}}_k) ||_F}{|| P_{\Omega}(\boldsymbol{\mathcal{M}})||_F} < \epsilon$ is achieved.
		\end{algorithmic}
	\end{algorithm}
	
	
	\subsection{Computational Complexity and Convergence Analysis}
	In this part, we first analyze the computational complexity of Algorithms~\ref{alg1}$\sim$\ref{alg3}. Since the TN topology updates dynamically in the optimization process, we only consider the general case, that is, the number of connections of each factor and the edge rank are $t$ and $R$, respectively. For an $N$-order incomplete tensor $\boldsymbol{{\mathcal{M}}}\in\mathbb{R}^{I_1\times I_2\times\cdots\times I_N} $ with $I_1=\cdots=I_N=I$, the computational cost of Algorithm~\ref{alg1} (ATL) on each iteration mainly includes two parts: 1) TN contraction in step 5; 2) updating $\boldsymbol{{\mathcal{Z}}}^{(i)} (i=1,\cdots, N)$ in step 6. The TN contraction aims to obtain $k$th low-rank component $\boldsymbol{\mathcal{A}}_k$ with tensor format, which conducts the matrix multiplication $N$-1 times and the computational complexity is $\mathcal{O}(NR^t I^{N})$. In step 6, the computational complexity of calculating $\boldsymbol{{\mathcal{Z}}}^{\neq n}\ $ and updating $\boldsymbol{{\mathcal{Z}}}^{(i)} (i=1,\cdots, N) $ are $\mathcal{O}(NR^t I^{N-1})$ and $\mathcal{O}(NR^{t}I^N+NR^{3t})$.
	Moreover, Algorithms~\ref{alg2} and \ref{alg1} have the same computational complexity due to the fact that the update cost of $\boldsymbol{{\mathcal {T}}}$ is negligible to the whole algorithm. Hence, the whole computational complexity of MTNR-ALS at each iteration is $\mathcal{O}(rNR^{t}I^N+rNR^{3t})$. In practice, such a heuristic algorithm requires more iterations due to the updating strategy. Compared with other TD-based methods including TR-ALS~\cite{ref37} and FCTN-PAM~\cite{ref39} with computational complexities costs being $\mathcal{O}(PNR^4 I^{N} + NR^{6})$ and $\mathcal{O}(N\sum_{k=2}^N I^kR^{k(N-k)+k-1}+NI^{N-1}R^{2(N-1)}+NR^{3(N-1)} )$, respectively, the overall computational complexity of MTNR-ALS is higher than TT-ALS, but it is lower than FCTN when facing higher-order tensors.
	
	For Algorithm~\ref{alg3} (MTNR-ADMM) with implicit low-rank regularization, the computational complexities of updating $\{\boldsymbol{{\mathcal{G}}}_{n}^{(i)}\}_{i=1,n=1}^{N,N}$ (step 7) is $\mathcal{O}(NIR^{t}(Rt+I))$. Then the computational complexities of Algorithm~\ref{alg3} is $\mathcal{O}(rN^2R^{t+1}I + rNR^t I^{N} + rNR^{3t})$. In comparison, for rank-minimization-based methods, SiLRTC~\cite{ref5} and TRNNM~\cite{ref61} require performing the full SVD in matricization format matrix and the total cost are $\mathcal{O}(NI^{N+1})$ and $\mathcal{O}(NI^{N+d})$ at each iteration. Therefore, imposing low-rank restraint in the latent factor can reduce considerable calculations.
	
	We establish the convergence guarantee of the proposed Algorithm~\ref{alg2}. Naturally, the convergence of Algorithms \ref{alg1} can also be obtained in a similar manner~\cite{wang2019global}. Since each low-rank component is designed to fit the current observation residuals, we just need to prove the convergence of the factors under a fixed structure.
	\begin{Theorem}(Sequence Convergence). 
		\label{th7}
		\it
		The sequence $\{\boldsymbol{\mathcal{Z}_k}^{(i)s} \}_{s\in \mathbb{N}} $ with $i=1,\cdots,N$ generated by algorithm \ref{alg2} converges to a critical point of (\ref{eq23}).
	\end{Theorem}
	The Lagrangian function of (\ref{eq23}) is $f(\{\boldsymbol{\mathcal{Z}_k}^{(i)} \})=||\mathcal{T} - \Re(\boldsymbol{\mathcal{Z}}^{(1)}_k,\cdots, \boldsymbol{\mathcal{Z}}^{(N)}_k)||_F^2+ \left \langle \boldsymbol{\mathcal{Q}}, P_{\Omega}(\boldsymbol{\mathcal{M}} - \sum_{i=1}^{k-1} \boldsymbol{\mathcal{A}}_i-\boldsymbol{\mathcal{T}}) \right \rangle$, where $\boldsymbol{\mathcal{Q}}$ is introduced as a Lagrangian multiplier. The solving strategy of model~(\ref{eq23}) is a special scenario in~\cite{xu2017globally}, so satisfying Theorem~\ref{th7} is based on the following four conditions.
	
	(a) The sequence of $f(\{\boldsymbol{\mathcal{Z}_k}^{(i)s} \})$ is non-increasing and $\{\boldsymbol{\mathcal{Z}_k}^{(i)s}\}\  (s\in \mathbb{N}) $ is a bounded sequence;
	
	(b) $\bigtriangledown f(\{\boldsymbol{\mathcal{Z}_k}^{(i)} \})$ has Lipschitz constant on any bounded set;
	
	(c) $\bigtriangledown_{\boldsymbol{\mathcal{Z}_k}^{(i)}} f(\{\boldsymbol{\mathcal{Z}_k}^{(i)s} \})$ has a Lipschitz constant $\alpha_i^s$ with respect to $\boldsymbol{\mathcal{Z}_k}^{(i)s} $, and there exists a constraint $0 < l \leq L < \infty$, such that $l\leq \alpha_i^s \leq L$ for all $i$ and $s$;
	
	(d) $f(\{\boldsymbol{\mathcal{Z}_k}^{(i)} \})$ satisfies the Kurdyka–Łojasiewicz property~\cite{bolte2007lojasiewicz} at $\{\boldsymbol{\mathcal{Z}_k}^{(i)s} \}_{s\in \mathbb{N}} $ .

	\begin{table}[]
		\centering
		\caption{The computational complexity of different methods on an $N$-order tensor of size $I\times \cdots\times I$. Here, $R$ is the rank of tensor decomposition models and $t$ is a constant.}
		\resizebox{.5\textwidth}{!}{
			\begin{tabular}{c|c}
				\toprule
				Method  &   Computational complexity\\
				\midrule
				TT-ALS~\cite{ref36}  & $\mathcal{O}((N-2)(PR^4I^N+R^6))$  \\
				TR-ALS~\cite{ref36}  & $\mathcal{O}(NPR^4I^N+NR^6)$  \\
				FCTN-PAM~\cite{ref39}  & $\mathcal{O}(N\sum_{k=2}^N I^kR^{k(N-k)+k-1}+NI^{N-1}R^{2(N-1)}+NR^{3(N-1)} )$\\
				TR-WOPT~\cite{ref62}  &  $\mathcal{O}(NR^2I^N+NR^4I^{N-1})$ \\
				MTNR-ALS  &  $\mathcal{O}(rNR^t I^{N} + rNR^{3t})$ \\
				\midrule
				SiLRTC~\cite{ref5}  & $\mathcal{O}(NI^{N+1})$  \\
				HaLRTC~\cite{ref5}  & $\mathcal{O}(NI^{N+1})$  \\
				TRNNM~\cite{ref61}  & $\mathcal{O}(NI^{N+d})$  \\
				TRLRF~\cite{ref44}  & $\mathcal{O}(NR^2I^N+NR^6)$  \\
				MTNR-ADMM  &  $\mathcal{O}(rNR^{t+1}It + rNR^t I^{N} + rNR^{3t})$ \\
				\bottomrule
		\end{tabular}}
		\label{tab1}
	\end{table}
	
	\begin{figure*}[ht]
		\centering
		\subfigure{
			\includegraphics[width=.24\textwidth]{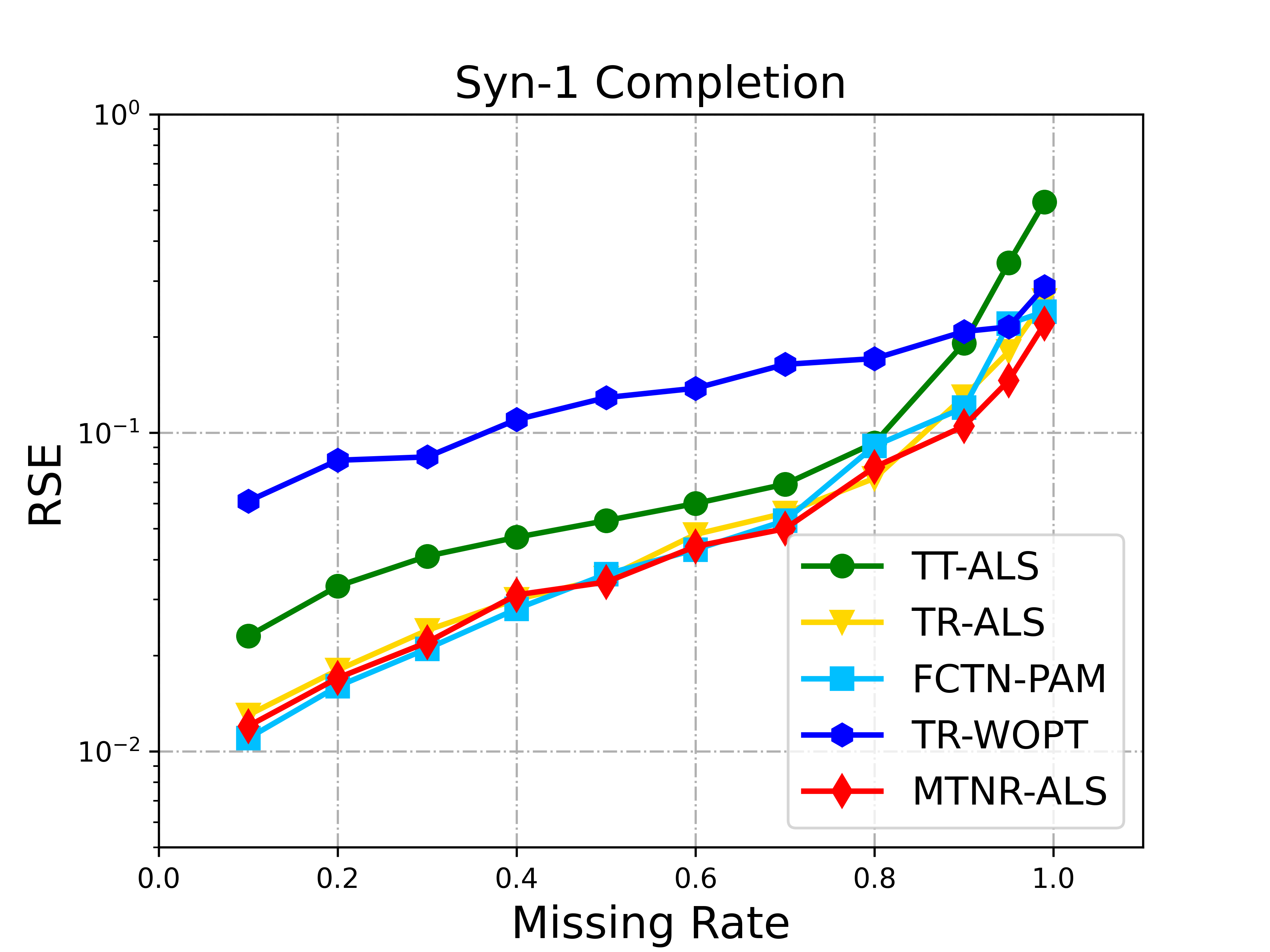} 
		}%
		\subfigure{
			\includegraphics[width=.24\textwidth]{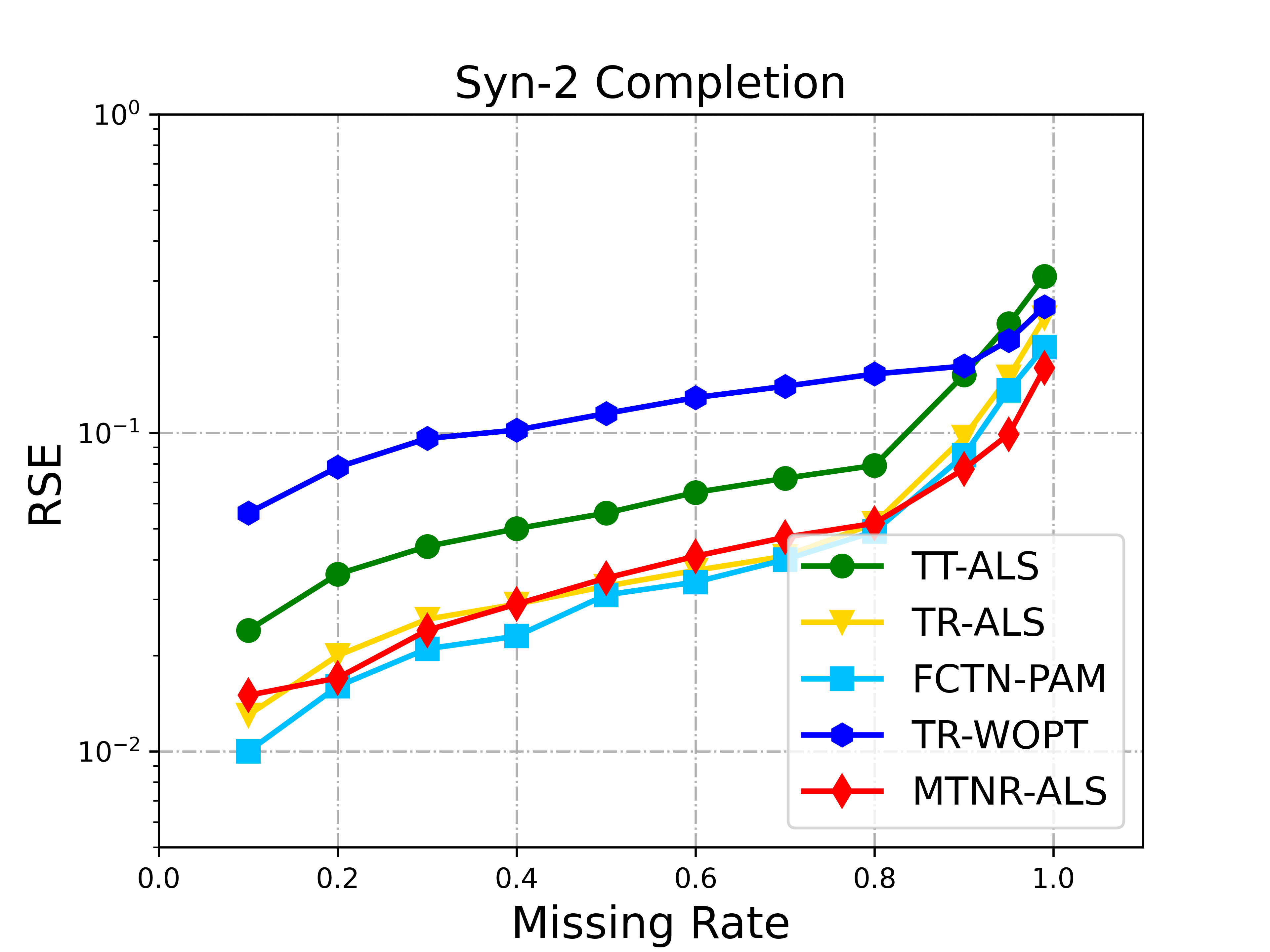} 
		}%
		\subfigure{
			\includegraphics[width=.24\textwidth]{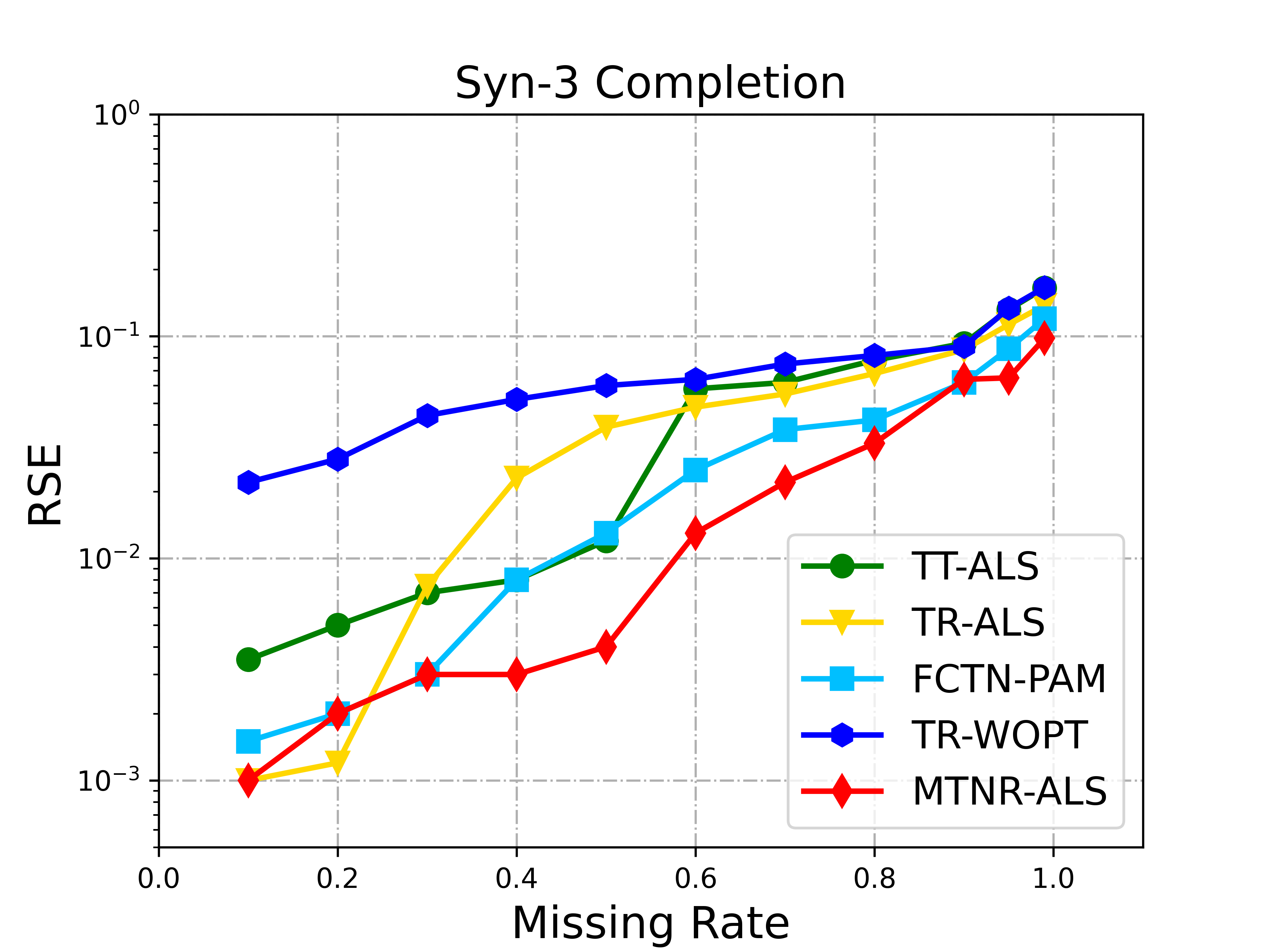} 
		}%
		\subfigure{
			\includegraphics[width=.24\textwidth]{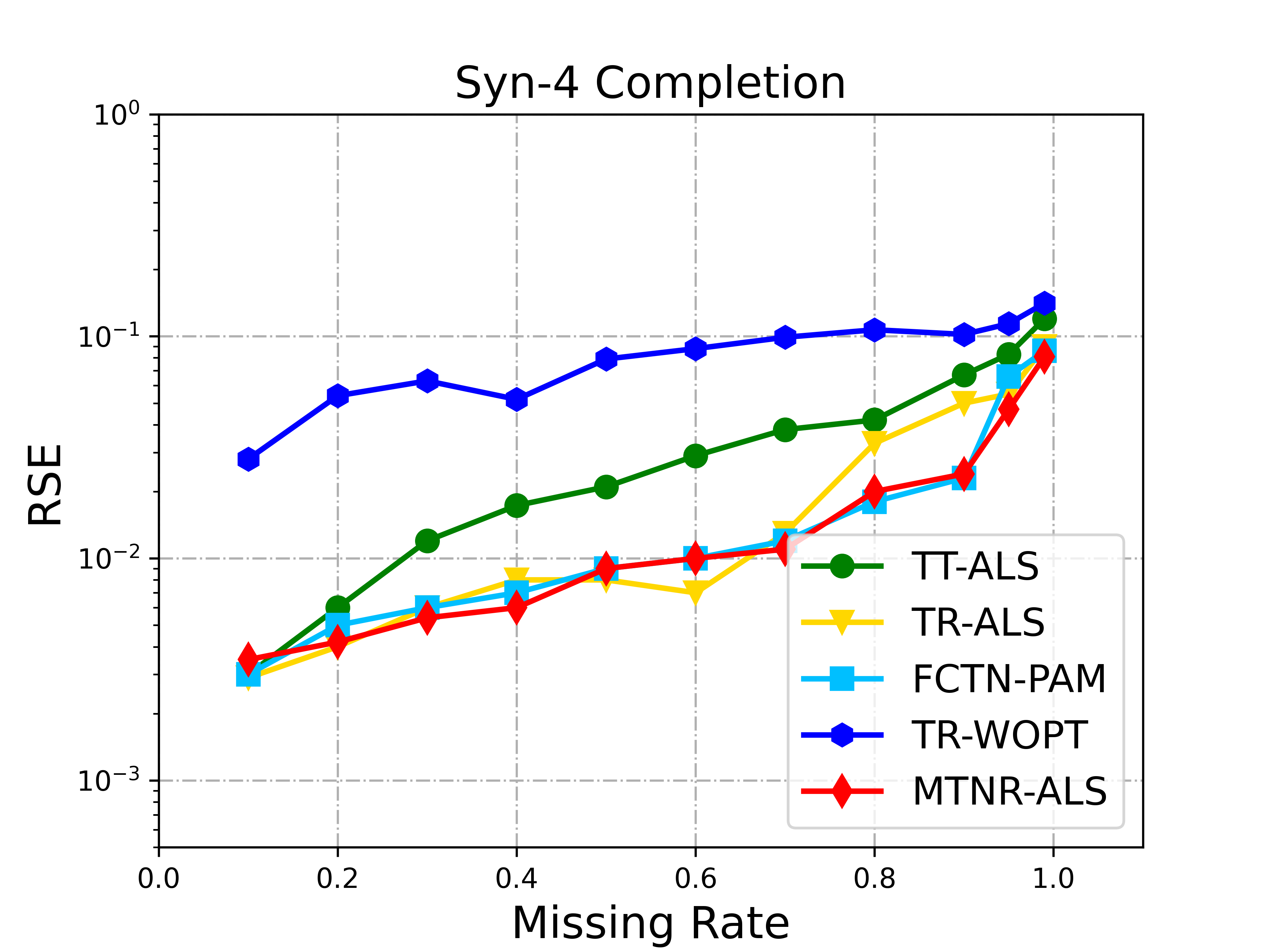} 
		}%
		\caption{Tensor completion performance of five TD-based methods on synthetic data (Syn-1$\sim$4) under different missing rates.}
		\label{img4}
	\end{figure*}

	\section{EXPERIMENTAL RESULTS}
	\label{sec5}
	In this section, we evaluate the proposed algorithms, MTNR-ALS and MTNR-ADMM, and compare them with other state-of-the-art TC methods. We first give the basic experimental settings. Then, extensive experimental results on synthetic data, benchmark color images, face datasets and color videos are presented to demonstrate the superiority and effectiveness of our approach.
	
	\subsection{Experimental Setting}
	\indent \textbf{1) Parameter setting and environment.}
	In our method, the two essential hyperparameters are the maximum number of connections $t$ and parameters upper bound $\gamma$, which control the computational and storage complexity of the algorithm. Unless otherwise stated, we set $t=3$ and $\gamma = NI4^t$, where $N$ and $I$ are the order and mode dimension of the instance tensor, respectively. Hence, the storage complexity of TN representation of low-rank components can be guaranteed to be linear with the data order. For Algorithms \ref{alg1}$\sim$\ref{alg3}, the relative error $\epsilon$, threshold $\delta$, and $S_{\max}$ are set to $2e$-2, $4e$-3, and $3e3$, respectively. For algorithm~\ref{alg3}, the trade-off coefficient $\lambda$,  positive penalty parameter $\rho$, and $\rho_{\max}$ are set to $1e$1, $1e$-1 and 30, respectively. All our experiments are performed on the same platform under Windows 10 on a PC with a 3.0 GHz CPU and 64G RAM.
	\vspace{2mm}
	
	\indent \textbf{2) Compared methods.}
	For comparison, we choose several start-of-the-art tensor completion algorithms, including TD-based TT-ALS~\cite{ref36}, TR-ALS~\cite{ref60}, FCTN-PAM~\cite{ref39}, and TR-WOPT~\cite{ref62}, where the FCTN-PAM corresponding TN topology can be regarded as a complete graph since it establishes connections among any two factors. The rank-minimization-based methods include SiLRTC~\cite{ref5}, HaLRTC~\cite{ref5}, and TRNNM~\cite{ref61}. The compare method based on both TD and rank-minimization is TRLRF~\cite{ref44}. We carefully adjust the parameters as suggested in the original paper for the above methods to obtain the best results. The computational complexity of the compared methods is shown in Table~\ref{tab1}.
	\vspace{2mm}

	\indent \textbf{3) Evaluation indices.}
	We adopt relative square error (RSE)~\cite{ref4}, peak signal-to-noise ratio (PSNR)~\cite{ref1_1}, and structure similarity (SSIM)~\cite{ref1_1} as quality indices to evaluate the performance of the reconstruction results. For synthetic data, RSE is applied as the quantitative evaluation metric and defined as
	\begin{equation}\begin{split}
			RSE = \frac{||\boldsymbol{\mathcal{X}}-\boldsymbol{\mathcal{Y}}||_F}{||\boldsymbol{\mathcal{Y}}||_F},
			\label{eq35}
	\end{split}\end{equation}
	where $\boldsymbol{{\mathcal{X}}}$ and $\boldsymbol{\mathcal{Y}} $ are the recovered tensor and ground truth, respectively. A smaller RSE value implies a better performance. For images data, PSNR and SSIM are applied as the quantitative evaluation metric and defined as
	\begin{equation}\begin{split}
			&PSNR = 10\log_{10}\frac{\boldsymbol{{Y}}_{\max}^2 \cdot num(\boldsymbol{{Y}})}{||\boldsymbol{{X}}-\boldsymbol{{Y}}||_F^2},\\
			& SSIM = \frac{(2\mu_{\boldsymbol{X}}\cdot \mu_{\boldsymbol{Y}} + c_1 ) (2\sigma_{\boldsymbol{X}\boldsymbol{Y}}+c_2)}{(\mu_{\boldsymbol{X}}^2+\mu_{\boldsymbol{Y}}^2+c_1)(\sigma_{\boldsymbol{X}}^2+\sigma_{\boldsymbol{Y}}^2+c_2)},
			\label{eq35}
	\end{split}\end{equation}
	where $\boldsymbol{X}$ and $\boldsymbol{Y} $ are the recovered image and ground truth image; $\mu_{\boldsymbol{X}}$ and $\sigma_{\boldsymbol{X}}$
	are the mean value and standard variance of $\boldsymbol{X}$; $\boldsymbol{Y}_{\max}$, $num(\boldsymbol{{Y}})$, $\mu_{\boldsymbol{Y}}$, and $\sigma_{\boldsymbol{Y}}$ are the maximum value, number of entries, mean value, and standard variance of $\boldsymbol{Y}$, respectively. The covariance of $\boldsymbol{X}$ and $\boldsymbol{Y} $ is $\sigma_{\boldsymbol{X}\boldsymbol{Y}}$, and $c_1,c_2>0$ are constant. Specifically, PSNR and SSIM measure the visual quality and structure similarity, and a large value of PSNR or SSIM indicates a better recovery accuracy.
	
	\begin{table*}[ht]
		\centering
		\caption{The RSE, PSNR, and SSIM values of the recovery results by ten different methods on ten color images under missing rate 90\%. The values that highlighted in bolder fonts are the best results.}
		\resizebox{1\textwidth}{!}{
			\begin{tabular}{c|c|ccccc|ccccc}
				\toprule
				
				Image                      & Metric & TT-ALS & TR-ALS & FCTN-PAM                  & TR-WOPT & MTNR-ALS & SiLRTC & HaLRTC & TRNNM  & TRLRF                     & MTNR-ADMM \\
				\midrule
				\multirow{3}{*}{Barbara}   & RSE    & 0.2272 & 0.1686 & 0.1738                  & 0.2884   & \textbf{0.1477}    & 0.3518 & 0.3025 & 0.2356 & 0.1644                     &\textbf{ 0.1489 }   \\
				& PSNR   & 19.31  & 21.89  & 21.63                    & 17.23   & \textbf{23.02}      & 15.50  & 16.81  & 18.98  & 22.11                      &\textbf{ 22.96 }    \\
				& SSIM   & 0.5498 & 0.6667 & 0.6455                  & 0.4286   & \textbf{0.7359}    & 0.4553 & 0.4991 & 0.6095 & 0.6786                     &\textbf{ 0.7351}    \\
				\midrule
				\multirow{3}{*}{House}    & RSE    & 0.1664 & 0.1061 & 0.1148                   & 0.1714  &\textbf{ 0.1028}    & 0.2591 & 0.2086 & 0.1518 & 0.0980                     & \textbf{0.09710 }  \\
				& PSNR   & 20.32  & 24.11  & 23.42                     & 19.98   &\textbf{ 24.37}    & 16.37  & 18.31  & 21.06  & 24.79                      & \textbf{24.88 }    \\
				& SSIM   & 0.5999 & 0.7348 & 0.6852                  & 0.5879     & \textbf{0.7506}  & 0.5931 & 0.6558 & 0.7556 & 0.7605                    &\textbf{ 0.7830}    \\
				\midrule
				\multirow{3}{*}{Peppers}   & RSE    & 0.2603 & \textbf{0.1789} & 0.1896                  & 0.2734   & 0.1811    & 0.3838 & 0.3576 & 0.2622 & 0.1723                   & \textbf{0.1691 }   \\
				& PSNR   & 18.17  &\textbf{ 21.00}  & 20.53                   & 17.36  & 20.95       & 14.67  & 15.28  & 17.95  & 21.32                      &\textbf{ 21.45}     \\
				& SSIM   & 0.5064 & \textbf{0.6303} & 0.5741                  & 0.4430   & 0.6301    & 0.4594 & 0.4830 & 0.5945 & 0.6498                  & \textbf{0.6778 }   \\
				\midrule
				\multirow{3}{*}{Sailboat} & RSE    & 0.2422 & 0.1832 & \textbf{0.1821}                    & 0.2383  & 0.1824   & 0.3284 & 0.3011 & 0.2254 & 0.1848                    & \textbf{0.1666 }   \\
				& PSNR   & 17.81  & 20.11  & 20.09                       & 17.80  & \textbf{20.39}    & 14.96  & 15.81  & 18.36  & 20.03                   & \textbf{20.83 }    \\
				& SSIM   & 0.4893 & 0.6013 & 0.5996                       & 0.4885 & \textbf{0.6120}    & 0.4531 & 0.4873 & 0.6065 & 0.5944                & \textbf{0.6555}    \\
				\midrule
				\multirow{3}{*}{Lena}      & RSE    & 0.2030 & \textbf{0.1610} & 0.1755                 & 0.2086    & 0.1628    & 0.2996 & 0.2627 & 0.2065 & 0.1544                   & \textbf{0.1499 }   \\
				& PSNR   & 19.15  & \textbf{21.07}  & 20.33                  & 18.84    & 20.96      & 15.80  & 16.94  & 18.96  & 21.41                     & \textbf{21.67}     \\
				& SSIM   & 0.5483 & 0.6244 & 0.5530                          & 0.5049 & \textbf{0.6312}    & 0.5045 & 0.5458 & 0.6343 & 0.6458                    & \textbf{0.6861}    \\
				\midrule
				\multirow{3}{*}{Baboon}   & RSE    & 0.2745 & 0.2582 & 0.2856                   & 0.2592   & \textbf{0.2499}   & 0.3473 & 0.3240 & 0.2656 & 0.2537                    & \textbf{0.2357}    \\
				& PSNR   & 16.58  & 17.11  & 16.22                   & 17.07   v  & \textbf{17.38}    & 14.53  & 15.15  & 16.86  &  17.25                  & \textbf{17.89}     \\
				& SSIM   & 0.3630 & 0.3793 & 0.3371                  & 0.3682    & \textbf{0.3902 }  & 0.3487 & 0.3591 & 0.4155 & 0.3834                  & \textbf{0.4338 }   \\
				\midrule
				\multirow{3}{*}{Airplane}  & RSE    & 0.1552 & 0.1150 & 0.1203                & 0.1865    & \textbf{0.1124}     & 0.2078 & 0.1832 & 0.1400 & 0.1197                   & \textbf{0.1058}    \\
				& PSNR   & 19.11  & 21.61  & 21.18                 & 17.35    & \textbf{21.79}       & 16.43  & 17.59  & 19.96  & 21.33                     & \textbf{22.25}     \\
				& SSIM   & 0.5721 & 0.6592 & 0.6440                & 0.4801     & \textbf{0.6943}    & 0.5318 & 0.5960 & 0.7026 & 0.6747                   & \textbf{0.7275}    \\
				\midrule
				\multirow{3}{*}{Facade}   & RSE    & 0.1507 & 0.1224 & 0.1482                 & 0.1942   & \textbf{0.1114}     & 0.2639 & 0.2015 & 0.1598 & 0.1188                    & \textbf{0.1174}    \\
				& PSNR   & 22.19  & 24.01  & 22.35                 & 19.99    & \textbf{24.82}       & 17.33  & 19.68  & 21.69  & 24.26                     & \textbf{24.36}     \\
				& SSIM   & 0.7044 & 0.7843 & 0.7193                & 0.5767    & \textbf{0.817}      & 0.4813 & 0.5732 & 0.7115 & 0.7913                   & \textbf{0.7948}    \\
				\midrule
				\multirow{3}{*}{Starfish}  & RSE    & 0.1216 & 0.0993 & 0.0925                & 0.1167    & \textbf{0.0833 }    & 0.1641 & 0.1252 & 0.0923 & 0.0907                    & \textbf{0.07422}   \\
				& PSNR   & 20.35  & 22.02  & 22.64                 & 20.65    & \textbf{23.53}       & 17.65  & 20.00  & 22.66  & 22.82                     & \textbf{24.54 }    \\
				& SSIM   & 0.6610 & 0.7374 & 0.7010                & 0.6369     & \textbf{0.7552}     & 0.6127 & 0.6994 & 0.7777 & 0.7457                  & \textbf{0.7927}    \\
				\midrule
				\multirow{3}{*}{Sea}       & RSE    & 0.1372 & 0.1049 & 0.1128                & 0.1509    & \textbf{0.1024}     & 0.2224 & 0.1713 & 0.1258 & \textbf{0.1021 }                   & 0.1029    \\
				& PSNR   & 21.11  & 23.45  & 22.81                 & 20.08     & \textbf{23.64 }      & 16.95  & 19.18  & 21.86  & 23.66                  & \textbf{24.37}     \\
				& SSIM   & 0.6769 & 0.7319 & 0.7220                & 0.6025    & \textbf{0.7455}     & 0.5848 & 0.6707 & 0.7329 & 0.7412                  & \textbf{0.7615}   \\
				\bottomrule
		\end{tabular}}
		\label{tab3}
	\end{table*}
	
	\begin{table}[]
		\centering
		\caption{Quantitative comparison of 5 TD-based methods on four synthetic data inpainting. The mode arrangement of each data includes original and transpositional and the sampling rate is 10\%.}
		\resizebox{.5\textwidth}{!}{
			\begin{tabular}{c|c|ccccc}
				\toprule
				Data  &   Permutation & TT-ALS & TR-ALS & FCTN-PAM                  & TR-WOPT & MTNR-ALS\\
				\midrule
				\multirow{2}{*}{Syn-1} & Original  & 0.192   &  0.137 & 0.125 & 0.202  & 0.108\\
				& Transpositional  &  0.224  & 0.149  & 0.125&  0.216  & 0.110\\
				\cline{1-7}
				\multirow{2}{*}{Syn-2} & Original  & 0.1524   & 0.0946  & 0.0853 & 0.162  & 0.0735\\
				& Transpositional  & 0.147   & 0.0901  & 0.0853& 0.155  & 0.0733\\
				\cline{1-7}
				\multirow{2}{*}{Syn-3} & Original  & 0.0969   & 0.082  & 0.0621 & 0.102  & 0.0640\\
				& Transpositional  & 0.0918   & 0.0784  &0.0620 & 0.096  &  0.0636\\
				\cline{1-7}
				\multirow{2}{*}{Syn-4} & Original  & 0.0750   & 0.0498  &0.0234 & 0.108  & 0.024\\
				& Transpositional  & 0.0628   & 0.0446  &0.023 &  0.0854  &0.022 \\
				\bottomrule
		\end{tabular}}
		\label{tab2}
	\end{table}

	\begin{figure}[h]
		\centering
		\subfigure[]{
			\includegraphics[width=.08\textwidth]{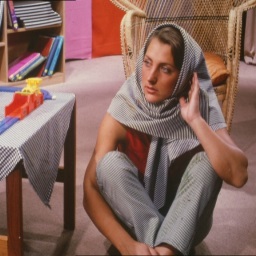} 
		}%
		\subfigure[]{
			\includegraphics[width=.08\textwidth]{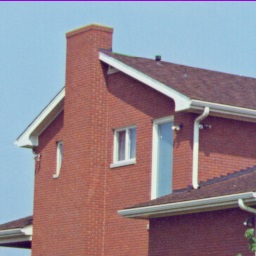} 
		}%
		\subfigure[]{
			\includegraphics[width=.08\textwidth]{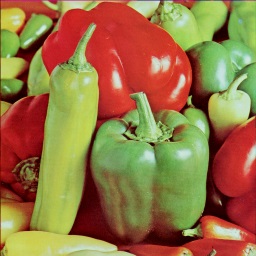} 
		}%
		\subfigure[]{
			\includegraphics[width=.08\textwidth]{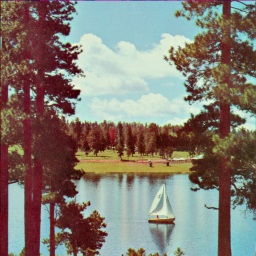} 
		}%
		\subfigure[]{
			\includegraphics[width=.08\textwidth]{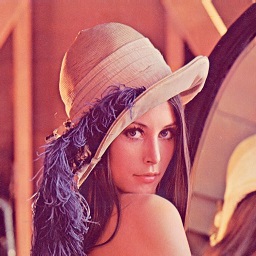} 
		}\\
		\subfigure[]{
			\includegraphics[width=.08\textwidth]{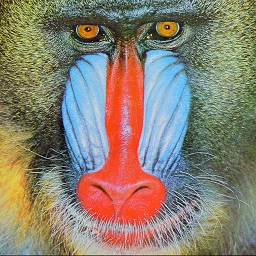} 
		}%
		\subfigure[]{
			\includegraphics[width=.08\textwidth]{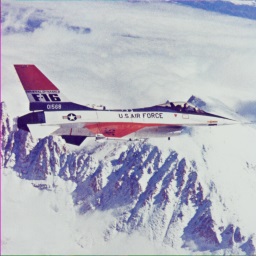} 
		}%
		\subfigure[]{
			\includegraphics[width=.08\textwidth]{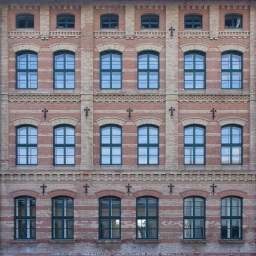} 
		}%
		\subfigure[]{
			\includegraphics[width=.08\textwidth]{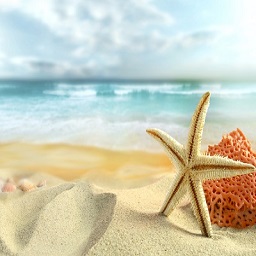} 
		}%
		\subfigure[]{
			\includegraphics[width=.08\textwidth]{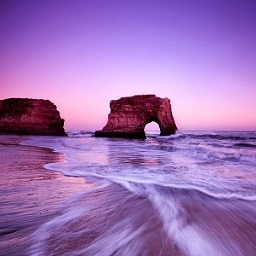} 
		}
		\caption{Ten original benchmark color images used for image completion, they are (a) Barbara, (b) House, (c) Peppers, (d) Sailboat, (e) Lena, (f) Baboon, (g) Airplane, (h) Facade, (i) Starfish, (j) Sea.}
		\label{img5}
	\end{figure}

	\subsection{Synthetic Data}
	
	To verify the robustness and effectiveness of the algorithms, we generated four different sorts of data. The Syn-1 is an ordinary $5$th-order tensor of size $8\times 8\times 8\times 8\times 8$, Syn-2 is a mode-unbalanced $5$th-order tensor of size $4 \times 8\times 12\times 16\times 20$, Syn-3 is a $5$th-order tensor of size $8\times 8\times 8\times 8\times 8$ with a large discrepancy in mode correlation, and Syn-4 is a $5$th-order tensor of size $4 \times 8\times 12\times 16\times 20$ with both mode imbalance and discrepancy. The Syn-1 and Syn-2 are obtained via the addition of 32 rank-1 tensors of the same size as the target tensor. The Syn-3 and Syn-4 are generated with TT representation of rank $(5,5,5,5)$ because TT only establishes connections within adjacent factors and has the problem of exponential decay of mode correlation~\cite{ref34}. Moreover, the elements of the generating factors are randomly sampled from the standard normal distribution $N(0,1)$. The compared methods include TT-ALS, TR-ALS, FCTN-PAW, TR-WOPT, and MTNR-ALS, where TR-WOPT and FCTN-PAW algorithms are based on gradient descent strategy and PAM framework~\cite{ref39}, respectively.
	
	\begin{figure*}[h]
		\centering
		\includegraphics[width=1.0\textwidth]{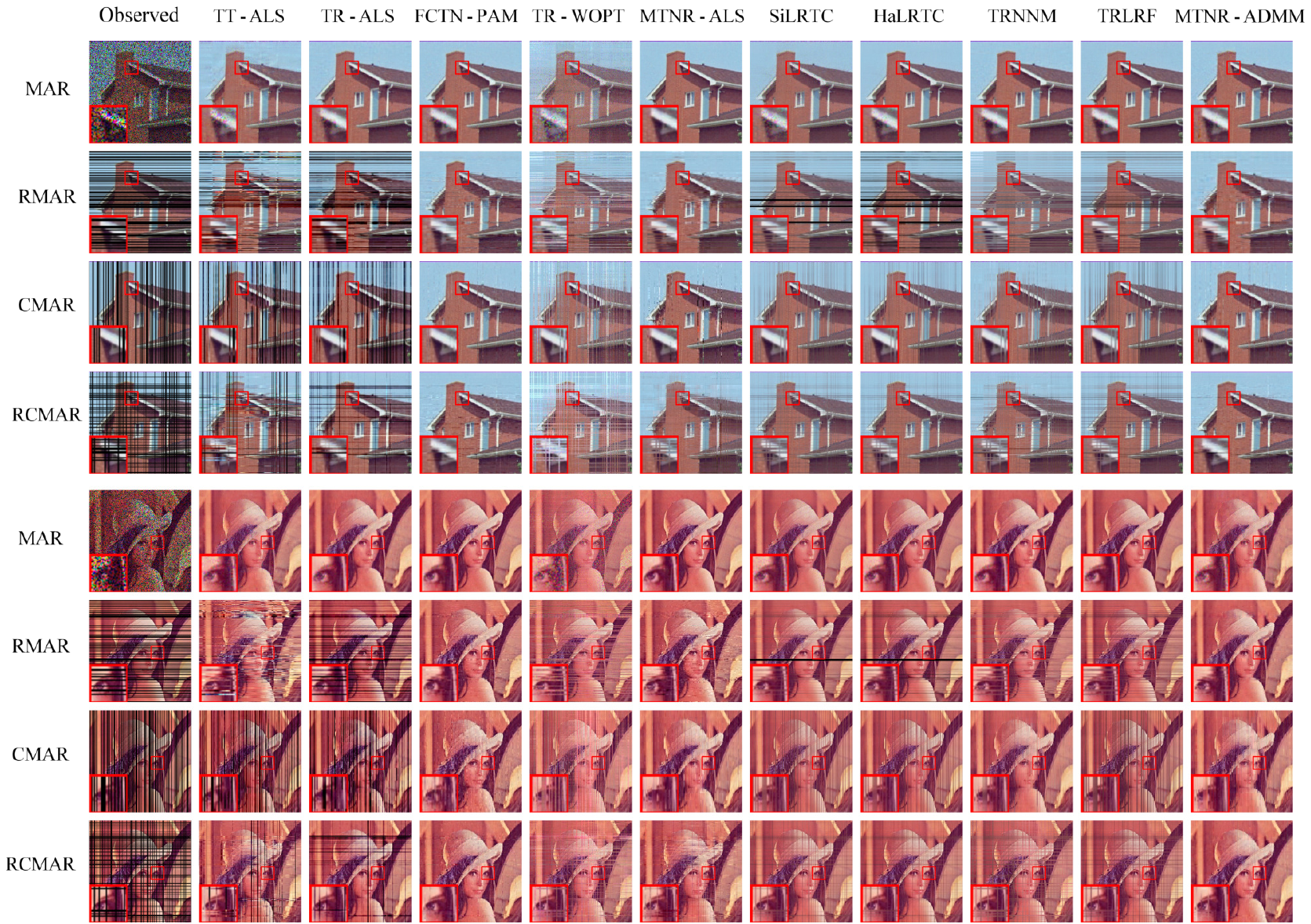} 
		\caption{The inpainting results of the image House and Lena by different methods under four missing patterns with a missing rate 40\%. From top to bottom, the missing patterns are MAR, RMAR, CMAR, and RCMAR, respectively. For better visualization, each image shows a magnified map of a local patch in the red box.}
		\label{img6}
	\end{figure*}
	
	\vspace{2mm}
	\textbf{1) Different missing rates.}
	We set the missing rate (MR) varying from 10\% to 99\%, and all entries are missing at random (MAR). In Fig.~\ref{img4}, we report the RSE values (best results of 10 independent experiments) of five TD-based methods. As can be seen, with the increase of MR, the completion performance of all algorithms decrease. Compared with other methods, MTNR-ALS achieves the best performance, and the advantage of MTNR-ALS is more apparent when the MR is larger than 80\%. Conversely, TR-WOPT yields the worst result and cannot obtain reliable convergence. Fig.~\ref{img4} shows that TR-WOPT cannot get a reasonable lower RSE value even under a lower MR (e.g., 10\%). In terms of time performance, MTNR-ALS requires more time than the counterparts since it requires updating structure additionally. It is worth noting that the ranks of the comparison methods, except MTNR-ALS, require being manually tuned to obtain the best results. In general, a smaller rank is recommended for an incomplete tensor with a large MR. However, the rank setting is challenging for high-dimensional data with different MR conditions, and it will significantly affect the accuracy of completion. Hence, a unique advantage of MTNR-ALS is not required to specify the rank in advance, and the rank of MTNR-ALS is generated automatically based on the rank incremental strategy. We noticed that MTNR-ALS delivered satisfactory results under different MR conditions, which attributes to its adaptive topology and rank. Compared with other TD models with fixed rank, our approach has better robustness and higher completion performance on synthetic data when MR is 99\%. This indicates the effectiveness of adaptive structural decomposition in TN completion based approaches.

	\vspace{2mm}

	\textbf{2) Different mode permutations.}
	As we know, TD models tend to sensitive to mode permutation, which is related to their corresponding topological structure. For instance, the ring-shaped TR model possesses circular dimensional permutation invariance~\cite{ref37}, but transposing adjacent modes of the original tensor will change the decomposition outcome. According to Section~\ref{sec3}, we know that the topology structure of MTNR is automatically generated during the optimization process and independent of the mode permutation. Therefore, MTNR has the property of transpositional invariance~\cite{ref39} as FCTN, which means the represented factors do not depend on the strict arrangement of the initial tensor mode. Notice that the factor contraction order is irrelevant to the results, and such property is produced during the tensor representation.
	
	We analyze the influence of the initial mode's arrangement of the incomplete tensor on the performance of the inpainting result by different algorithms. We transpose the 1st and 4th modes of Syn-1, the 2nd and 5th modes of Syn-2, the 2nd and 4th modes of Syn-3, and the 1st and 5th modes of Syn-4, respectively. The quantitative RSE values of inpainting results by different methods on four synthetic data (original and transpositional) are listed in Table~\ref{tab2}, where the MR is 90\%. From Table~\ref{tab2}, the recovery results of TT and TR based approaches vary significantly under different mode arrangements. Fortunately, the MTNR and FCTN based approaches are robust to tensor mode transposition and obtain better results. This further provides empirical evidence for the mode arrangement invariance of MTNR.


	\begin{table*}[h]
		\centering
		\caption{The RSE, PSNR, and SSIM values of the recovery results by ten different methods on images House and Lena under different missing patterns with a missing rate 40\%. The values highlighted in bolder fonts are the best results.}
		\resizebox{1\textwidth}{!}{
			\begin{tabular}{c|c|c|ccccc|ccccc}
				\toprule
				
				Image                   & \multicolumn{1}{l}{Missing pattern} & Metric & TT-ALS & TR-ALS & FCTN-PAM & TR-WOPT & MTNR-ALS  & SiLRTC & HaLRTC & TRNNM   & TRLRF  & MTNR-ADMM \\
				\midrule
				\multirow{12}{*}{House} & \multirow{3}{*}{MAR}        & RSE    & 0.0826 & 0.0517 & 0.0474   & 0.1513  & \textbf{0.03108}  & 0.0702 & 0.0441 & 0.0442  & 0.0509 & 0.0488    \\
				&                             & PSNR   & 26.33  & 30.34  & 31.10    & 21,06   & \textbf{35.64}    & 27.72  & 31.74  & 31.52   & 30.45  & 30.84     \\
				&                             & SSIM   & 0.8502 & 0.9228 & 0.9313   & 0.5989  & \textbf{96.56}    & 0.8984 & 0.9566 & 0.9534  & 0.9250 & 0.9202    \\
				\cline{2-13}
				& \multirow{3}{*}{RMAR}       & RSE    & 0.5046 & 0.4851 & {0.0974}   & 0.2194  & 0.09984  & 0.2372 & 0.2872 & 0.1940  & 0.1743 & \textbf{0.0928}    \\
				&                             & PSNR   & 10.76  & 11.01  & {24.93 }   & 17.82   & 24.76    & 17.12  & 15.26  & 18.89   & 19.86  & \textbf{25.39}     \\
				&                             & SSIM   & 0.3018 & 0.2708 & {0.8379}   & 0.4788  & 0.8506   & 0.6585 & 0.5036 & 0.6253  & 0.6438 & \textbf{0.8617}    \\
				\cline{2-13}
				& \multirow{3}{*}{CMAR}       & RSE    & 0.5387 & 0.4435 & \textbf{0.0649}   & 0.1812  & 0.1121   & 0.1163 & 0.1099 & 0.1476  & 0.1076 & 0.0693    \\
				&                             & PSNR   & 10.10  & 11.74  & \textbf{28.95}    & 19.48   & 23.62    & 23.38  & 23.90  & 21.29   & 24.05  & 27.81     \\
				&                             & SSIM   & 0.2456 & 0.2930 & \textbf{0.8980}    & 0.5429  & 0.8087  & 0.7900 & 0.8070 & 0.7131  & 0.8258 & 0.8862    \\
				\cline{2-13}
				& \multirow{3}{*}{RCMAR}      & RSE    & 0.4552 & 0.3008 & {0.0729}   & 0.1718  & 0.1146   & 0.1016 & 0.0857 & 0.1276  & 0.0958 & \textbf{0.0686}    \\
				&                             & PSNR   & 11.47  & 15.21  & {27.40}    & 19.98   & 23.52    & 24.52  & 26.01  & 22.60   & 25.18  & \textbf{27.92 }    \\
				&                             & SSIM   & 0.3006 & 0.5226 & {0.8503}   & 0.5627  & 0.8101   & 0.8127 & 0.8519 & 0.7527  & 0.8260 & \textbf{0.8825}    \\
				\midrule
				\multirow{12}{*}{Lena}  & \multirow{3}{*}{MAR}        & RSE    & 0.1040 & 0.0814 & 0.0628   & 0.1901  & \textbf{0.0537}  & 0.0904 & 0.0741 & 0.06224 & 0.0802 & 0.0679    \\
				&                             & PSNR   & 24.87  & 26.93  & 29.18    & 19.73   &\textbf{ 30.53}    & 26.08  & 27.79  & 29.27   & 27.06  & 28.50     \\
				&                             & SSIM   & 0.8382 & 0.8901 & 0.9232   & 0.6286  & \textbf{0.9328}   & 0.8781 & 0.9181 & 0.9360  & 0.8923 & 0.9176    \\
				\cline{2-13}
				& \multirow{3}{*}{RMAR}       & RSE    & 0.3474 & 0.4807 & 0.09132  & 0.1264   & 0.1683 & 0.2024  & 0.1863 & 0.1611  & 0.1246 & \textbf{0.0867 }   \\
				&                             & PSNR   & 14.50  & 11.86  & 25.88    & 23.15    & 20.86  & 19.20   & 20.06  & 21.16   & 23.31  & \textbf{26.40}     \\
				&                             & SSIM   & 0.4905 & 0.2812 & 0.8496   & 0.7458   & 0.8011 & 0.6094  & 0.7477 & 0.7099  & 0.8036 & \textbf{0.8871}    \\
				\cline{2-13}
				& \multirow{3}{*}{CMAR}       & RSE    & 0.5872 & 0.5282 & 0.1253    & \textbf{0.1159}  & 0.1464  & 0.1797 & 0.1785 & 0.2366  & 0.1169 & 0.1253    \\
				&                             & PSNR   & 9.87   & 10.89  & 23.26    & \textbf{23.90}   & 21.94    & 20.13  & 20.17  & 17.83   & 23.83  & 23.83     \\
				&                             & SSIM   & 0.1952 & 0.2521 & 0.7414   & 0.7913  & 0.7257   & 0.6629 & 0.6638 & 0.5758  & \textbf{0.8138} & 0.7785    \\
				\cline{2-13}
				& \multirow{3}{*}{RCMAR}      & RSE    & 0.3446 & 0.3134 & 0.0927   & 0.1904  & 0.1489   & 0.1367 & 0.1289 & 0.1617  & 0.1711 & \textbf{0.0892}    \\
				&                             & PSNR   & 14.53  & 15.52  & 25.83    & 19.65   & 21.72    & 22.51  & 22.97  & 21.09   & 20.64  & \textbf{26.16}     \\
				&                             & SSIM   & 0.5040 & 0.5526 & 0.84907   & 0.6304  & 0.7213  & 0.7629 & 0.7791 & 0.7034  & 0.6875 & \textbf{0.8687}    \\
				\bottomrule
		\end{tabular}}
		\label{tab4}
	\end{table*}
	
	\begin{table*}[ht]
		\centering
		\caption{Comparison of average RSE, PSNR, AND SSIM for different algorithms recovered results on four face data (Data-1$\sim$Data-4 ) with a missing rate 70\%. Each data includes 64 face images phoned under different illumination conditions. The values that highlighted in bolder fonts are the best results.}
		\resizebox{1\textwidth}{!}{
			\begin{tabular}{c|c|ccccc|ccccc}
				\toprule
				
				Data                      & Metric & TT-ALS & TR-ALS & FCTN-PAM                  & TR-WOPT & MTNR-ALS & SiLRTC & HaLRTC & TRNNM  & TRLRF                     & MTNR-ADMM \\
				\midrule
				\multirow{3}{*}{Data-1} & RSE  & 0.2347          & 0.2503 & 0.2582   & 0.2804   & 0.2462  & 0.4010 & 0.2540          & 0.2909          & 0.2306          & \textbf{0.2257} \\
				& PSNR & 29.09           & 28.45  & 28.28    & 27.53    & 28.66   & 24.26  & 28.62           & 28.16           & 29.48           & \textbf{29.61}  \\
				& SSIM & \textbf{0.9015} & 0.8691 & 0.8556   & 0.8387   & 0.8768  & 0.8425 & 0.8969          & 0.8994          & 0.8924          & 0.8919          \\
				\midrule
				\multirow{3}{*}{Data-2} & RSE  & 0.1940          & 0.1766 & 0.1998   & 0.2831   & 0.1779  & 0.4199 & 0.2144          & 0.2504          & \textbf{0.1718} & 0.1728          \\
				& PSNR & 27.10           & 27.89  & 26.93    & 24.13    & 27.89   & 22.13  & 26.61           & 25.75           & 28.28           & \textbf{28.33}  \\
				& SSIM & 0.8932          & 0.9052 & 0.8864   & 0.8098   & 0.8991  & 0.8500 & 0.9106          & 0.9074          & 0.9140          & \textbf{0.9150} \\
				\midrule
				\multirow{3}{*}{Data-3} & RSE  & 0.2054          & 0.1839 & 0.2184   & 0.3015   & 0.1809  & 0.4049 & 0.2149          & 0.2569          & 0.1884          & \textbf{0.1755} \\
				& PSNR & 26.34           & 27.31  & 25.90    & 23.14    & 26.90   & 21.99  & 26.38           & 25.29           & 27.26           & \textbf{27.69}  \\
				& SSIM & 0.8798          & 0.8870 & 0.8606   & 0.7876   & 0.8854  & 0.8394 & \textbf{0.9102} & 0.8996          & 0.8971          & 0.8936          \\
				\midrule
				\multirow{3}{*}{Data-4} & RSE  & 0.2503          & 0.2653 & 0.2638   & 0.3187   & 0.2408  & 0.4635 & 0.2805          & 0.3196          & 0.2600          & \textbf{0.2396} \\
				& PSNR & 24.84           & 24.33  & 24.44    & 22.84    & 25.09   & 21.22  & 24.48           & 23.77           & 25.03           & \textbf{25.36}  \\
				& SSIM & 0.8540          & 0.8361 & 0.8274   & 0.7773   & 0.8516  & 0.8200 & 0.8862          & \textbf{0.8835} & 0.8642          & 0.8644 \\
				\bottomrule
		\end{tabular}}
		\label{tab5}
	\end{table*}
	
	\begin{figure*}[h]
		\centering
		\includegraphics[width=1.0\textwidth]{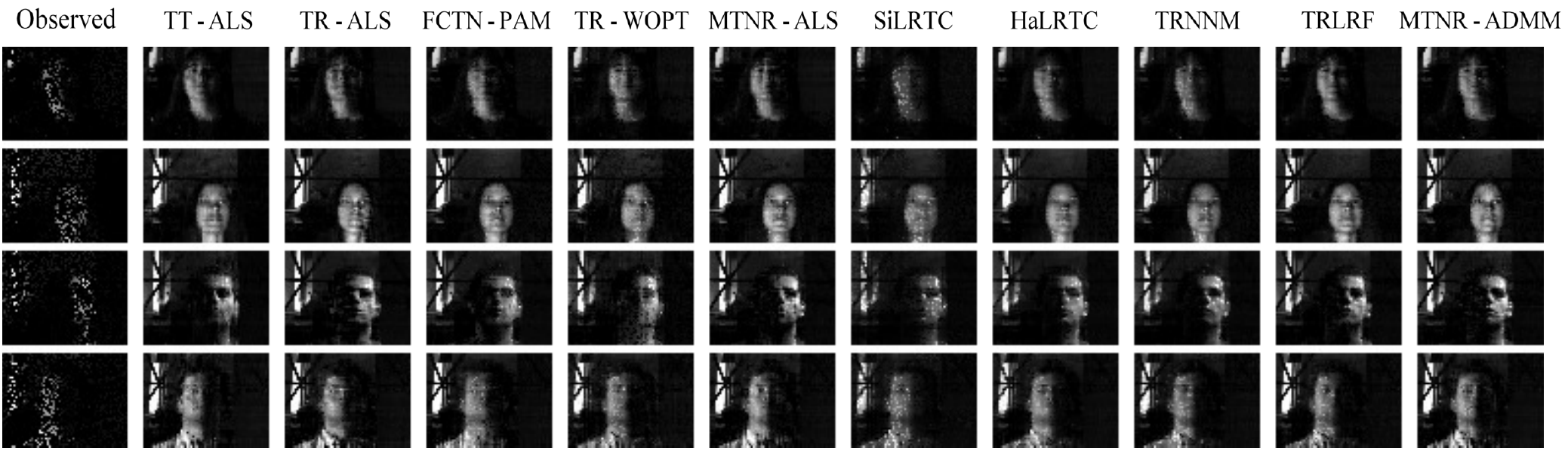} 
		\caption{Four examples of recovered grayscale face images for different algorithms with a missing rate 70\%. From top to bottom, these are the first images of Data-1$\sim$Data-4 of size $48\times 64$. }
		\label{img7}
	\end{figure*}
	
	\begin{figure*}[h]
		\centering
		\includegraphics[width=1.0\textwidth]{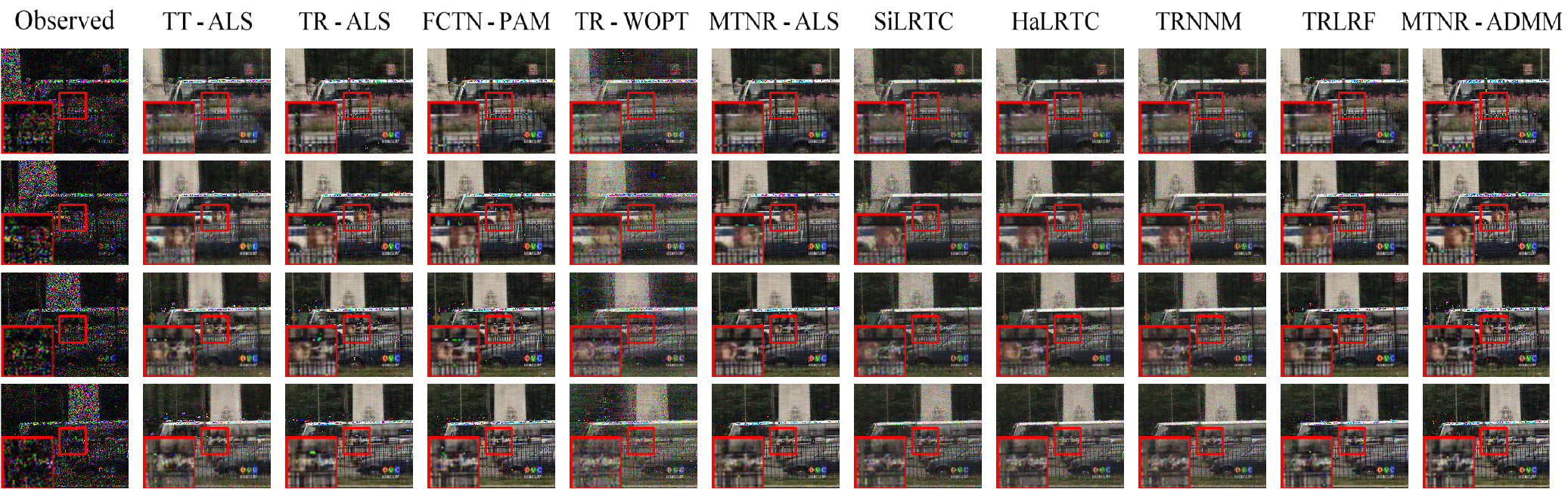} 
		\caption{Visual comparison of video completion by different algorithms on Bus video sequences. From top to bottom, there are the 16th, 32th, 48th, and 64th frames in the video, respectively.}
		\label{img8}
	\end{figure*}
	
	\begin{figure*}[h]
		\centering
		\subfigure[]{
			\includegraphics[width=1\textwidth]{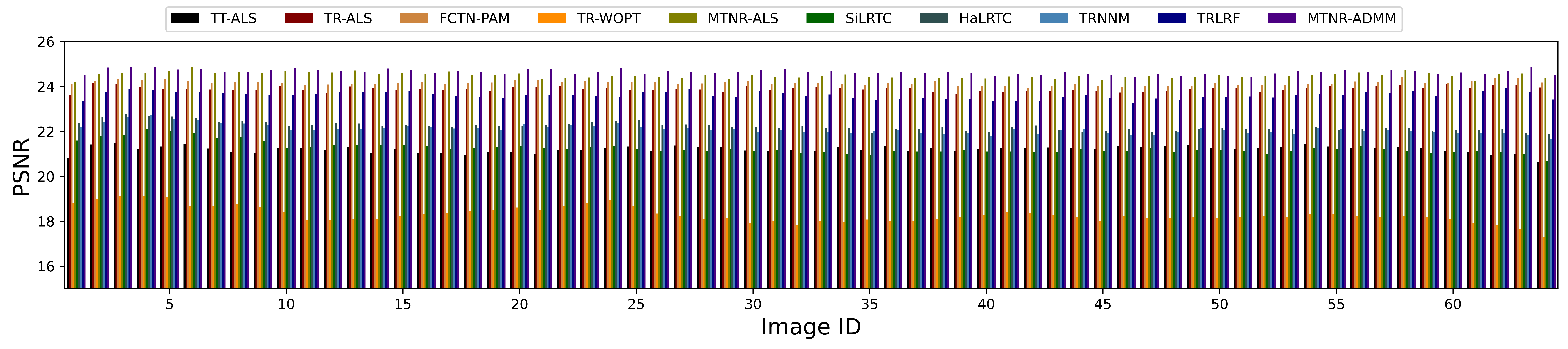} 
		}\\
		\subfigure[]{
			\includegraphics[width=1\textwidth]{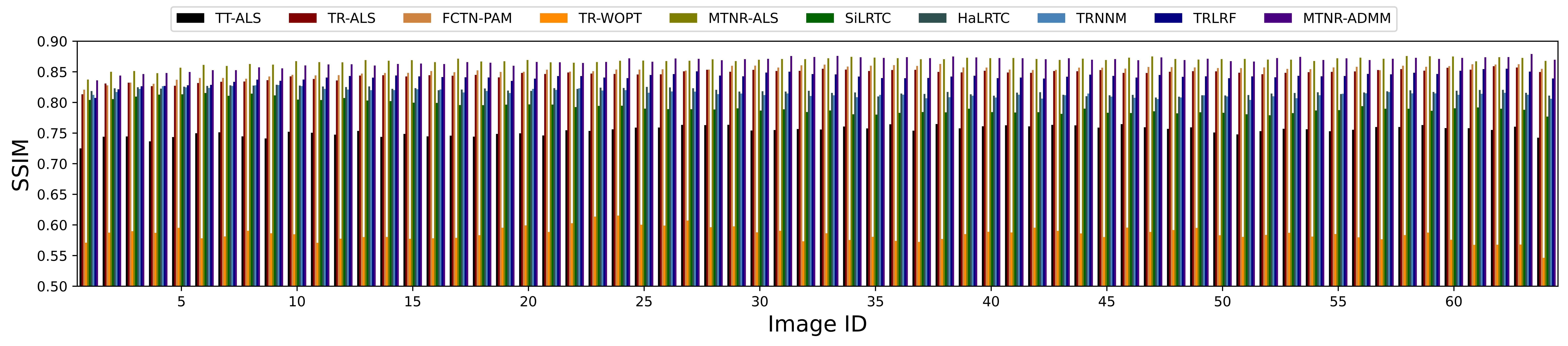} 
		}
		\caption{Comparison of the PSNR and SSIM values for different algorithms on 64 images of the test Bus video with a missing rate 60\%.}
		\label{img9}
	\end{figure*}

\subsection{Image Completion}
	We further evaluate the performance of MTNR-ALS and MTNR-ADMM in the color image completion task, and compare them with eight other start-of-the-art algorithms. Ten benchmark color images of size $256\times 256\times 3$ are shown in Fig.~\ref{img5}, which are reshaped to 5th-order tensor of size $16\times 16\times16\times 16\times 3$ in the experimental process.
	
	\vspace{2mm}
	\textbf{1) Completion with uniformly random missing.}
	We use 10\% randomly sampled entries (with a MR of 0.9) to perform the image completion task. For comparison methods, the parameter settings follow their original paper to achieve the best result. The methods TR-ALS, TR-WOPT, and TRLRF, represent the target tensor with TR format and the rank is set as 8. The rank of TT-ALS is a vector $(8, 12, 12, 6)$. In addition, the rank of FCTN-PAM is set to 4 due to limited computing resources. The quantitative indices RSE, PSNR, and SSIM are used and the best inpainting results among twenty independent experiments are reported in Table~\ref{tab3}. Note that the overall recovery performance is unsatisfactory due to the local structural information loss during the reshaping process and a higher MR. 
	
	For the TD-based methods in the left hand side of Table~\ref{tab3}, MTNR-ALS achieved the best completion performance on most of the test examples. Many results show FCTN-PAM is a little inferior to TR-ALS due to its limited robustness to target tensors with unbalanced mode dimensions. Besides, for higher-order tensors ($N>5$), the optimization of FCTN-PAM is expensive since its computational and storage complexity increases exponentially with the tensor order. We find the gradient descent-based method TR-WOPT unstable during the optimization process and obtain poor reconstruction accuracy. For the TC methods with rank-minimization regularization in the right half of Table~\ref{tab3}, the SiLRTC, HaLRTC, and TRNNM directly impose conditions on the target tensor and obtain large RSE and small PSNR and SSIM values. The TRLRF and MTNR-ADMM with latent restraints achieve better performance than their counterparts TR-ALS and MTNR-ALS, demonstrating the performance gains due to TD and rank-minimization. Finally, our proposed MTNR-ADMM achieves the best performance in natural image completion tasks with uniformly random missing.
	
	\vspace{2mm}
	\textbf{2) Completion with nonrandom missing.}
	We also consider the image completion in the case of Row-wise missing at random (RMAR), Colum-wise missing at random (CMAR), and Row-Column-wise missing at random (RCMAR). The images "House" and "Lena" were selected for non-randomly missing completion experiments. We report the impainting image in Fig.~\ref{img6} and report the quantitative indices in Table.~\ref{tab4} by ten different methods under different miss patterns with a miss rate of 40\%. Each image in Fig.~\ref{img6} shows a magnified map of a local patch (in the red box) for clarity. 
	
	We observe that the TD-based algorithms with ALS solver hardly deal with nonrandom missing cases, e.g., TT-ALS and TR-ALS. Surprisingly, the gradually generated TD model can greatly alleviate this problem and obtain acceptable results. Although the inpainting performance of MTNR-ALS is inferior to FCTN, which is optimized under the PAM framework, it is far superior to TR-ALS. The SiLRTC and HaLRTC have similar poor recovery performance in nonrandom missing cases, which implies the limitations of SNN. Most of the other methods have a relatively apparent residual stripe in the visualization results, while MTNR-ADMM has better visual effects. The quantitative indices of MTNR-ADMM in Table~\ref{tab4} also achieve the most satisfactory outcomes, especially in the RMAR pattern of the second image, and the SSIM value is 0.0375 higher than the second one. These experiments suggest that MTNR-ADMM can still work excellently under other unrandom missing patterns.
	
	\subsection{YaleFace Dataset Completion}
	In this section, we verify the performance of the proposed algorithm on Extended Yale Face Database B. The test data include the 5th, 10th, 15th, and 20th people in the first pose under 64 illumination conditions, and each image was resized to $48\times 64$. Hence, we get four sets of data (Data-1$\sim$4)with a size of $48\times64\times64$, and in the experiment, we reshape them to $48\times64\times8\times8$ and randomly select $p\times$100\% pixel values as the observation data. We set the observation level $p=0.3$, and $t$ and $\gamma$ of MTNR are set to 2 and $NI8^2$, respectively. Note that when $t=2$, MTNR is also easy to form a ring topology, but the ranks of different edges and mode arrangements are dynamically adjusted. The completion task for human faces is difficult since the features under different illuminations are more complicated, and the downsampled images have a lower local similarity.
	
	We report the average completion performance of each data by different algorithms in Table~\ref{tab5}. Unlike the previous experimental results, we found that the inpainting results of FCTN-PAM are inferior to TT-ALS and TR-ALS. Two reasons are considered: 1) Excessive connections within factors will lead to over-parameterization of TD models, which is unfavourable for complicated data representation; 2) The identical edge rank will cause the inflexible representation since a small rank change will lead to a more significant storage cost adjustment. That implies that we don't need to establish a connection between any factors, and the informative features of data's different modes can be propagated along the edge. From the partial visualization results in Fig.~\ref{img7}, the global structure of the inpainting results of TRLRF and MTNR-ADMM are more intact.
	
	\subsection{Video Completion}
	We also evaluate the proposed two algorithms on color video completion tasks and the test video sequence "Bus" comes from the YUV dataset~\cite{ref11}. We selected the first 64 frames of the "Bus" video and each frame with size $144\times176\times3$. Then, the size of the test tensor is $144\times176\times3\times64$. Compared with the face data in the previous subsection, the adjacent frames of the video sequence are highly similar, and there is more redundant information. During the experiment, the ranks of TT, TR and FCTN based approaches are set as 20, 18, and 8, respectively. We focus on comparing MTNR and FCTN to highlight the advantage of the ATL algorithm on the rank determination of TD models. The $t$ and $\gamma$ of MTNR are equivalent to 3 and $NI8^t$. We randomly selected 40\% of pixels as observation entries for the video completion task. We visualize the inpainting results of the 16th, 32th, 48th, and 64th frames in Fig.~\ref{img8} and compare each frame's PSNR and SSIM values in Fig.~\ref{img9}. From the visualization of video completion, the edges of the recovery images of SiLRTC and TR-WOPT are discontinuous and blurry. At the same time, MTNR-ADMM delivers the best completion results, and the inpainting images are clearer than the others. Notably, we only generated one low-rank component for MTNR in the experimental process, which corresponds to an FCTN model with ranks (16,3,12;16,3,13;3,3,3). Empirical evidence shows that the rank determination based on the ATL algorithm is more effective than the simple manual setting.
	
	\section{Conclusion}
	\label{sec6}
	
	This paper proposed a generalized MTNR framework, which expresses a high-order tensor as a superposition of multiple low-rank components in a TN format. The topology of MTNR models is data-adaptive and robust to the mode arrangement of the data. We defined the basic multilinear operations of the tensor with TN format in theory and developed a novel ATL algorithm for latent factors learning of MTNR in practice. Then, two algorithms, called MTNR-ALS and MTNR-ADMM, are presented for the LTRC task. Extensive experimental results on synthetic data, benchmark color image, face dataset, and video demonstrate that the performance of MTNR is outperforming the state-of-the-art TD and rank-minimization based methods.

	\section*{Acknowledgment}
	Postgraduate Research \& Practice Innovation Program (KYCX21\_0304) of Jiangsu Province, China. The authors would like to thank Andong Wang for his suggestions to improve this paper.
	

	\appendices
	%
	%
	%
	%
	%
	

	
	
	%
	

	\bibliographystyle{IEEEtran}
	\bibliography{IEEEexample}
	
	%
	
	
	
	
	
	
	

\end{document}